\documentclass{article} 
\usepackage{iclr2027_conference,times}

\usepackage{amsmath,amsfonts,bm}

\def\eqref#1{equation~\ref{#1}}

\def\1{\bm{1}}

\DeclareMathAlphabet{\mathsfit}{\encodingdefault}{\sfdefault}{m}{sl}
\SetMathAlphabet{\mathsfit}{bold}{\encodingdefault}{\sfdefault}{bx}{n}

\usepackage{hyperref}
\usepackage{url}

\usepackage{enumitem}
\usepackage{booktabs}
\usepackage{multirow}
\usepackage{adjustbox}
\usepackage{graphicx}
\usepackage{capt-of}
\usepackage{needspace}
\usepackage{amssymb}
\usepackage[table]{xcolor}

\newcommand{\ours}{\textsc{SAT-ACT}}
\newcommand{\oursbase}{\textsc{SAT-ACT (Pref. Only)}}

\newcommand{\pancell}[1]{\cellcolor{yellow!40}#1}

\title{SAT-ACT: Learning CDCL Branching from Outcome-Derived Action Preferences}

\author{Zhenchao Sun, Chongyang Tao, Ping Lu, Tongkai Lu, Jianling Gao, Shuai Ma \\
SKLCCSE Lab, Beihang University, Beijing, China \\
\texttt{\{zhenchao.sun,chongyang,luping\}@buaa.edu.cn} \\
\texttt{\{lutongkai,jianlingg,mashuai\}@buaa.edu.cn}
}

\iclrfinalcopy

\begin{document}

\maketitle

\begin{abstract}

Learning-based approaches have increasingly been explored for SAT solving, with particular attention to improving branching decisions in conflict-driven clause learning (CDCL) solvers.
These approaches derive branching guidance either from static structural properties of SAT formulas or from dynamic signals obtained through solver feedback and execution traces.
However, these forms of supervision provide limited evidence about state-dependent branching effects, while different outcome measures may favor different branching choices.
To address these limitations, we propose \textsc{SAT-ACT}, an outcome-derived learning framework for CDCL branching.
We design three complementary outcome comparison criteria to extract action preferences from independent branching action rollouts initiated at the same solver state.
The criteria identify preferences supported by the outcomes and leave comparisons unresolved when the evidence is insufficient or conflicting, instead of forcing an ordering using a predefined scalar metric.
A state-conditioned network learns branching priorities from the extracted preferences and uses them to guide CDCL search directly without online action rollouts.
We further introduce non-dominated heuristic action supervision, which retains the solver-selected action as a training target only when it remains non-dominated under these preferences.
Solver-level experiments show that \textsc{SAT-ACT} consistently reduces CDCL search costs on three SAT families and improves PAR-2 on the SAT Competition benchmarks.
\end{abstract}

\section{Introduction}
\label{sec-intro}
 
The Boolean satisfiability problem (SAT) is a fundamental problem in computer science with broad applications in planning~\citep{aips/ButtnerR05}, verification~\citep{pieee/VizelWM15}, and security~\citep{sat/MironovZ06}.
As the first problem shown to be NP-complete~\citep{stoc/Cook71}, SAT has motivated decades of research on efficient solving algorithms.
Modern SAT solvers are largely built upon conflict-driven clause learning (CDCL), which combines branching, Boolean constraint propagation, conflict analysis, clause learning, and restart mechanisms to solve large and complex instances~\citep{handbook2021}.
Among these components, branching is particularly consequential because each branching action selects an unassigned variable together with its polarity, and this choice can substantially alter the subsequent search trajectory.
Modern CDCL solvers therefore rely on carefully designed branching heuristics to select these actions.

Recently, many studies have explored learning-based approaches to solve SAT instances or enhance SAT solving~\citep{corr/BunzL17,neurosat_SelsamLBLMD19,iclr/AmizadehMW19,dac/LiSLKCX23,aaai/CameronCHL20,neurocore_SelsamB19,satformer_iccad23,NeuroBack_iclr24,sun2026unsat}.
A representative line of work learns structural signals from SAT formulas to guide CDCL heuristics.
Such methods predict information related to unsatisfiable cores or backbone variables to assist variable selection and phase initialization~\citep{neurocore_SelsamB19,satformer_iccad23,NeuroBack_iclr24,sun2026unsat}.
More recent approaches instead derive supervision from the solving process itself.
Reinforcement-learning (RL) methods learn branching guidance through solver interaction or solver-level performance feedback~\citep{nips/KurinGWC20,rlaf/abs-2505-16053}, while imitation-based methods learn branching policies from decisions recorded in CDCL execution traces~\citep{zhang2026boolean}.
Together, these developments move learning-based branching from static formula-level guidance toward supervision that more directly reflects dynamic CDCL search.

Although existing learning-based approaches have improved CDCL branching, they construct supervision at different levels of the solving process.
Methods that learn structural signals from SAT formulas guide variable selection and phase initialization, but these signals do not describe the downstream effect of a branching decision at a dynamic solver state~\citep{neurocore_SelsamB19,satformer_iccad23,NeuroBack_iclr24,sun2026unsat}.
RL-based methods learn from solver feedback~\citep{nips/KurinGWC20,rlaf/abs-2505-16053}, while imitation-based methods learn from decisions recorded during CDCL execution~\citep{zhang2026boolean}.
These forms of supervision reflect solver behavior, but they do not necessarily provide matched outcomes for multiple branching decisions at the same dynamic state.
The effect of a branching decision depends on the current assignments, learned clauses, and search history, so evidence collected from different states or trajectories cannot provide matched comparisons at a particular decision point.
Classical SAT lookahead solvers probe candidate branches through propagation-based effects~\citep{biere2021look,handbook2021}, while strong branching in mixed-integer linear programming (MILP) evaluates candidate variables under a shared branch-and-bound context~\citep{khalil2016learning,gasse2019exact}.
These precedents demonstrate the value of assessing candidate decisions under a shared search context.
Existing learning-based CDCL branching methods, however, still provide limited supervision for distinguishing the downstream effects of the branching decisions available at a dynamic solver state.

A further challenge lies in deriving reliable supervision from solver outcomes at a dynamic solver state.
Solver outcomes report whether solving terminates and how many conflicts, propagations, decisions, and restarts occur during the evaluation.
These measures describe different aspects of the subsequent search and may favor different branching choices.
Reducing them to a single scalar criterion would impose a fixed tradeoff among different outcome measures.
The key question is therefore when the observed outcomes provide sufficient evidence to prefer one branching choice over another.
A suitable supervision scheme should retain only preferences supported by the observed outcomes and leave comparisons with insufficient or conflicting evidence unordered rather than forcing a complete ordering over the available choices.

To address these limitations, we propose \ours{}, a framework for learning CDCL branching from outcome-derived action preferences.
The framework samples candidate branching actions at selected solver states and runs an independent CDCL rollout for each action from its corresponding state.
The rollout outcomes are examined using three complementary outcome comparison criteria to extract supported action preferences.
To learn branching decisions from these preferences, we develop a CDCL branching action network that encodes the active CNF and dynamic assignment features and produces state-conditioned scores for branching actions.
The extracted preferences supervise the network through a pairwise ranking objective.
During solving, the learned scores guide branching directly without requiring action rollouts.
To incorporate guidance from the solver's internal branching heuristic, we introduce non-dominated heuristic action supervision.
Under this mechanism, the heuristic action is used as a training target only when it remains non-dominated under the extracted preferences.

Our main contributions are summarized below.

\begin{itemize}[
  topsep=2pt,
  partopsep=0pt,
  itemsep=2.5pt,
  parsep=0pt
]

\item We introduce an outcome-derived learning framework for CDCL branching.
The framework uses a state-conditioned model to score branching actions, while outcome-derived action preferences supervise the model through a pairwise ranking objective.
During solving, the trained model guides branching directly without online
action rollouts.

\item We design three complementary outcome comparison criteria for extracting supported preferences between branching actions.
These criteria leave action pairs without sufficient evidence unordered and
avoid collapsing the observed outcomes into a predefined metric.

\item We additionally propose non-dominated heuristic action supervision for the action selected by the solver's internal branching heuristic.
The heuristic action is retained as a training target only when it remains non-dominated under the outcome-derived preferences.

\end{itemize}

\section{Preliminaries}
\label{sec-pre}

\paragraph{Boolean Satisfiability Problem (SAT).}

We consider a Boolean formula $\phi$ in conjunctive normal form (CNF) over $N$ Boolean variables and $M$ clauses~\citep{handbook2021}.
Each variable $v_i$ takes either \textit{true} or \textit{false}, represented by $1$ and $0$, respectively.
A CNF formula is a conjunction of clauses, and each clause is a disjunction of literals.
A literal $\ell$ is either a variable $v_i$ or its negation $\lnot v_i$.
These two literals are referred to as the positive and negative literals of $v_i$ and have opposite polarities.
We use
$\mathcal{L}=\{\ell_0,\ldots,\ell_{2N-1}\}$
to denote the set of all literals and
$\mathcal{C}=\{c_0,\ldots,c_{M-1}\}$
to denote the set of clauses in $\phi$.
The SAT problem asks whether there exists an assignment under which $\phi$ evaluates to \textit{true}.
A formula admitting such an assignment is \textit{satisfiable}.
Otherwise, it is \textit{unsatisfiable}.

\paragraph{Conflict-Driven Clause Learning (CDCL).}

CDCL forms the basis of modern SAT solvers~\citep{handbook2021}.
During solving, CDCL repeatedly performs Boolean constraint propagation and branching decisions, while conflicts trigger conflict analysis, clause learning, and non-chronological backtracking.
Let $S_t$ denote the dynamic solver state immediately before the $t$-th branching decision.
The state contains the current partial assignment, dynamic search information, and the active clause set $\mathcal{C}_t$.
The set $\mathcal{C}_t$ contains the original and learned clauses retained by the solver at $S_t$.

\paragraph{CDCL Branching Actions.}
At each branching decision, CDCL determines an unassigned variable and its polarity.
These choices are commonly made by separate variable-selection and
phase-selection heuristics.
For example, Exponential Variable State-Independent Decaying Sum (EVSIDS)
maintains activity scores that emphasize recent conflict participation and
selects the eligible unassigned variable with the highest activity~\citep{handbook2021}.
A phase-selection heuristic then determines its polarity.
We represent these choices jointly as a decision literal, referred to as the
\emph{branching action}.
Applying the action assigns the selected literal to true.
Let $\mathcal{B}_t\subseteq\mathcal{L}$ denote the set of legal branching
actions at $S_t$.
For each eligible unassigned variable, $\mathcal{B}_t$ contains its positive
and negative literals.
We denote the action selected by the solver's internal branching heuristic as
$a_t^{\mathrm{heur}}\in\mathcal{B}_t$.
Different branching actions from the same solver state can produce different
propagation effects, search trajectories, and solving outcomes.

\section{Method}
\label{sec:method}

\begin{figure*}[t]
    \centering
    \includegraphics[width=\textwidth]{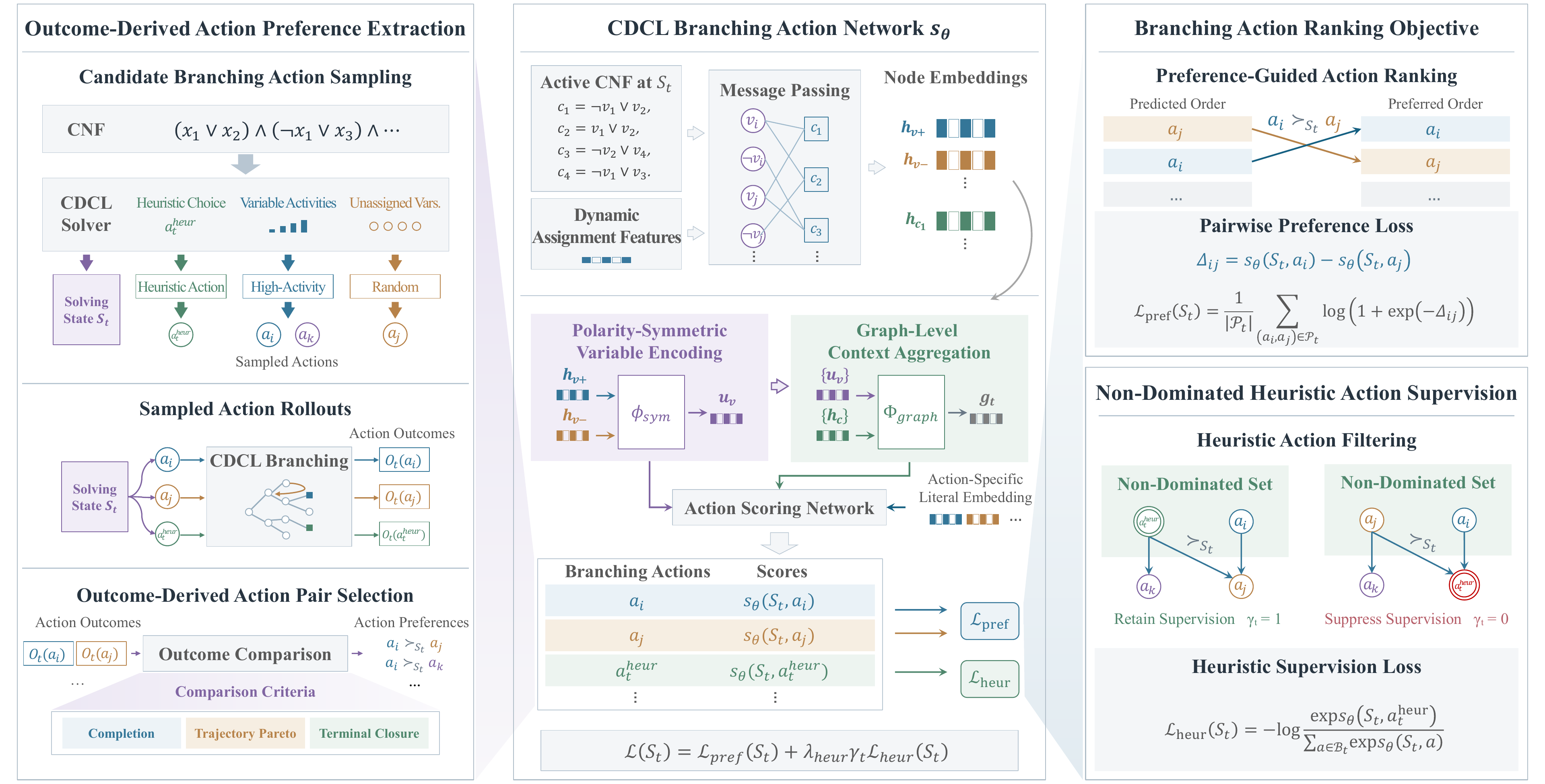}
    \vspace{-20px}
    \caption{
Overview of \ours{}, an outcome-derived learning framework for CDCL branching.
Candidate branching actions initiate independent CDCL rollouts from the same solver state.
The rollout outcomes are compared to extract action preferences, which supervise a state-conditioned branching action network through pairwise ranking.
    }
    \label{fig:satact_framework}
    \vspace{-10px}
\end{figure*}

In this section, we present \ours{}, an outcome-derived learning framework for CDCL branching.
At selected solver states, the framework samples candidate branching actions and runs an independent CDCL rollout for each action from its corresponding state.
We compare the rollout outcomes using three complementary criteria to extract supported action preferences, while uncertain comparisons remain unordered.
A CDCL branching action network encodes the active CNF and dynamic assignment features to produce state-conditioned scores for branching actions.
The extracted preferences supervise the network through a pairwise ranking objective.
We also introduce non-dominated heuristic action supervision for the action selected by the solver's internal branching heuristic.
The heuristic action is retained as a training target only when it remains non-dominated under the extracted preferences.
Figure~\ref{fig:satact_framework} presents an overview of the framework.

\subsection{Outcome-Derived Action Preference Extraction}
\label{subsec:outcome_preferences}

CDCL branching is state-dependent, while static supervision or a single observed trajectory provides limited evidence about other legal branching actions available at the same point in the search.
We therefore run independent CDCL rollouts for multiple candidate branching actions from the same dynamic solver state and record their solving outcomes.
Because all rollouts begin from the same state, their outcomes can be compared under a shared search context.
We then apply three complementary outcome comparison criteria to extract supported action preferences while leaving uncertain comparisons unordered.

We first sample solver states along CDCL search trajectories.
For each sampled state $S_t$, we construct a bounded candidate set
$\mathcal{A}_t\subseteq\mathcal{B}_t$.
The heuristic action $a_t^{\mathrm{heur}}$ is retained in $\mathcal{A}_t$.
Each action $a\in\mathcal{A}_t$ initiates an independent CDCL rollout from $S_t$.
The action is applied as the next branching decision, after which the solver resumes its internal CDCL procedure under a common configuration and rollout budget.
Each rollout produces an outcome
\begin{equation}
    O_t(a)
    =
    \bigl(
        y_t(a),
        C_t(a),
        P_t(a),
        D_t(a),
        R_t(a)
    \bigr),
\end{equation}
where $y_t(a)\in\{\mathrm{SAT},\mathrm{UNSAT},\mathrm{incomplete}\}$ denotes the rollout status, and
$C_t(a)$, $P_t(a)$, $D_t(a)$, and $R_t(a)$ record the numbers of conflicts, propagations, decisions, and restarts.
The state sampling, candidate construction, and rollout protocol are detailed in Appendix~\ref{app:outcome_preference_details}.
These rollouts are used only for training-data collection and are not required during inference.

For two actions $a_i,a_j\in\mathcal{A}_t$, we write
$a_i\succ_{S_t}a_j$
when their rollout outcomes provide sufficient evidence for preferring $a_i$ to $a_j$.
The resulting preference set is
\begin{equation}
    \mathcal{P}_t
    =
    \{(a_i,a_j)\mid a_i\succ_{S_t}a_j\}.
\end{equation}
Because the observed outcomes may not distinguish every action pair, $\mathcal{P}_t$ defines a partial preference relation.
Rather than combining the outcome measures into a predefined scalar score, we use three complementary comparison criteria to determine which preferences are supported.

\paragraph{Criterion 1. Completion (C1).}

Let $\tau_t(a)=\mathbb{I}[y_t(a)\in\{\mathrm{SAT},\mathrm{UNSAT}\}]$ indicate whether the rollout of action $a$ reaches a terminal result within the rollout budget.
C1 supports $a_i\succ_{S_t}a_j$ when $\tau_t(a_i)>\tau_t(a_j)$.
Thus, an action whose rollout reaches SAT or UNSAT is preferred to one whose rollout remains incomplete.
If both rollouts have the same completion status, C1 provides no preference evidence.

\paragraph{Criterion 2. Trajectory Pareto (C2).}

For each completed rollout, we define the search-cost vector
$\mathbf{c}_t(a)=(C_t(a),P_t(a),D_t(a),R_t(a))$.
C2 supports $a_i\succ_{S_t}a_j$ when
$\mathbf{c}_t(a_i)\preceq\mathbf{c}_t(a_j)$ component-wise and
$\mathbf{c}_t(a_i)\neq\mathbf{c}_t(a_j)$.
This criterion prefers an action only when its rollout requires no more conflicts, propagations, decisions, or restarts and improves at least one of these quantities.
If the two rollouts exhibit trade-offs across the search-cost dimensions, C2 provides no preference evidence.

\paragraph{Criterion 3. Terminal Closure (C3).}

For each completed rollout, we define a terminal-closure level $\kappa_t(a)$ that measures how directly the rollout reaches its final SAT or UNSAT result.
A larger value indicates that less additional search is required after applying the branching action.
The exact terminal-closure levels are defined in Appendix~\ref{app:terminal_closure}.
C3 supports $a_i\succ_{S_t}a_j$ when $\kappa_t(a_i)>\kappa_t(a_j)$.
Equal terminal-closure levels provide no preference evidence.

For each action pair, we combine the evidence provided by the applicable comparison criteria.
The pair $(a_i,a_j)$ is added to $\mathcal{P}_t$ only when at least one criterion supports $a_i\succ_{S_t}a_j$ and no criterion supports the opposite direction.
If no criterion distinguishes the actions or the available evidence is conflicting, the pair remains unordered.

\subsection{Learning CDCL Branching from Action Preferences}
\label{subsec:learning_cdcl_branching}

After extracting outcome-derived action preferences, we use them to train a state-conditioned CDCL branching action network.
At solver state $S_t$, the network encodes the active CNF and dynamic
assignment features to score each legal branching action $a\in\mathcal{B}_t$
with $s_\theta(S_t,a)$, which determines its relative ranking.

\paragraph{Literal--Clause Graph-Based CDCL State Representation.}

We represent the active CNF at $S_t$ as a literal--clause bipartite graph
$\mathcal{G}_t=(\mathcal{L}_t,\mathcal{C}_t,\mathcal{E}_t)$.
Let $\mathcal{V}_t$ denote the variables represented in the active CNF, and let
$\mathcal{L}_t=\{v^{+},v^{-}\mid v\in\mathcal{V}_t\}$
contain their positive and negative literals.
The set $\mathcal{C}_t$ contains the active clauses, and an edge
$(\ell,c)\in\mathcal{E}_t$
indicates that literal $\ell$ occurs in clause $c$.

Literal-node features encode the current assignment status and assignment level together with occurrence and global search information.
Clause-node features encode the clause size, the current assignment composition of its literals, and its satisfaction and near-unit status.
Detailed definitions of these features are provided in Appendix~\ref{app:state_features}.

The literal and clause features are projected into a shared embedding space to obtain initial representations
$\mathbf{h}^{(0)}_{\ell}$ and $\mathbf{h}^{(0)}_{c}$.
We then apply a NeuroCore-style message-passing encoder~\citep{neurocore_SelsamB19}.
Literal and clause nodes exchange information along their incidence edges, while each literal also incorporates information from its opposite-polarity counterpart.
After several rounds, the encoder produces contextual literal and clause embeddings
$\mathbf{h}_{\ell}$ and $\mathbf{h}_{c}$.

\paragraph{Polarity-Aware Branching Action Scoring.}

For each variable $v$, let $\mathbf{h}_{v^{+}}$ and
$\mathbf{h}_{v^{-}}$ be its positive and negative literal embeddings.
Polarity-symmetric variable encoding combines them as
\begin{equation}
    \mathbf{u}_v
    =
    \phi_{\mathrm{sym}}
    \left(
        \left[
            \mathbf{h}_{v^{+}}+\mathbf{h}_{v^{-}}
            \,;\,
            \left|
                \mathbf{h}_{v^{+}}-\mathbf{h}_{v^{-}}
            \right|
        \right]
    \right),
\end{equation}
where $\phi_{\mathrm{sym}}$ is a shared learnable transformation.
The resulting representation is invariant to exchanging the two polarities,
while $\mathbf{h}_{v^{+}}$ and $\mathbf{h}_{v^{-}}$ retain
polarity-specific information.

Graph-level context aggregation summarizes the variable and clause
representations as
\begin{equation}
    \mathbf{g}_t
    =
    \Phi_{\mathrm{graph}}
    \left(
        \{\mathbf{u}_v\}_{v\in\mathcal{V}_t},
        \{\mathbf{h}_c\}_{c\in\mathcal{C}_t}
    \right).
\end{equation}
The graph-level representation $\mathbf{g}_t$ captures the global context of
the current solver state.

Graph-conditioned variable encoding then incorporates this context into each
variable representation
\begin{equation}
    \widetilde{\mathbf{u}}_v
    =
    \Phi_{\mathrm{var}}
    \left(
        \mathbf{u}_v,
        \mathbf{g}_t
    \right),
\end{equation}
where $\Phi_{\mathrm{graph}}$ and $\Phi_{\mathrm{var}}$ are learnable
aggregation and conditioning functions.
Their implementations are detailed in
Appendix~\ref{app:action_scoring}.

For a branching action $a$ associated with variable $v(a)$, the network
computes
\begin{equation}
    s_\theta(S_t,a)
    =
    \phi_{\mathrm{score}}
    \left(
        \left[
            \mathbf{h}_a
            \,;\,
            \widetilde{\mathbf{u}}_{v(a)}
            \,;\,
            \mathbf{g}_t
        \right]
    \right).
\end{equation}
The action-specific literal embedding $\mathbf{h}_a$ distinguishes the two
polarities of the same variable, while
$\widetilde{\mathbf{u}}_{v(a)}$ and $\mathbf{g}_t$ provide variable-level
and graph-level context.

\paragraph{Preference-Guided Action Ranking.}

The preference set $\mathcal{P}_t$ supervises the relative ordering of the
branching-action scores.
For each $(a_i,a_j)\in\mathcal{P}_t$ with
$a_i\succ_{S_t}a_j$, we define the score difference
\begin{equation}
    \Delta_{ij}
    =
    s_\theta(S_t,a_i)-s_\theta(S_t,a_j).
\end{equation}
We impose the extracted preference relations through a logistic pairwise
ranking objective
\begin{equation}
    \mathcal{L}_{\mathrm{pref}}(S_t)
    =
    \frac{1}{|\mathcal{P}_t|}
    \sum_{(a_i,a_j)\in\mathcal{P}_t}
    \log\left(1+\exp(-\Delta_{ij})\right).
\end{equation}
This objective encourages the model to assign higher scores to preferred
actions without requiring absolute action-quality targets.
The loss is averaged within each state so that states with different numbers
of supported preference pairs contribute on a comparable scale.
We set $\mathcal{L}_{\mathrm{pref}}(S_t)=0$ when
$\mathcal{P}_t$ is empty.

\paragraph{Non-Dominated Heuristic Action Supervision.}

The action selected by the solver's internal branching heuristic can provide
a training target, but it may conflict with the outcome-derived preferences
when an evaluated action is preferred to it.
We therefore retain this supervision only when the heuristic action remains
non-dominated among the evaluated candidates.

Let $\mathcal{N}_t$ denote the non-dominated subset of
$\mathcal{A}_t$
\begin{equation}
    \mathcal{N}_t
    =
    \left\{
        a\in\mathcal{A}_t
        \;\middle|\;
        \nexists\,a'\in\mathcal{A}_t
        \text{ such that }
        a'\succ_{S_t}a
    \right\}.
\end{equation}
Since $\succ_{S_t}$ is a partial preference relation and
$\mathcal{A}_t\subseteq\mathcal{B}_t$, membership in
$\mathcal{N}_t$ does not imply global optimality.
It only indicates that no evaluated action is preferred to that action.

Let $a_t^{\mathrm{heur}}$ denote the heuristic action.
We define the supervision gate as
\begin{equation}
    \gamma_t
    =
    \mathbb{I}
    \left[
        a_t^{\mathrm{heur}}\in\mathcal{N}_t
    \right].
\end{equation}
When the heuristic action remains non-dominated under the outcome-derived
preferences, we use it as a training target through a cross-entropy objective
over the legal branching actions
\begin{equation}
    \mathcal{L}_{\mathrm{heur}}(S_t)
    =
    -\log
    \frac{
        \exp s_\theta(S_t,a_t^{\mathrm{heur}})
    }{
        \sum_{a\in\mathcal{B}_t}
        \exp s_\theta(S_t,a)
    }.
\end{equation}

\paragraph{Overall Training Objective.}

The state-level training objective combines preference-guided action ranking
with non-dominated heuristic action supervision
\begin{equation}
    \mathcal{L}(S_t)
    =
    \mathcal{L}_{\mathrm{pref}}(S_t)
    +
    \lambda_{\mathrm{heur}}
    \gamma_t
    \mathcal{L}_{\mathrm{heur}}(S_t),
\end{equation}
where $\lambda_{\mathrm{heur}}$ controls the contribution of heuristic action
supervision.

Together, the two terms train the network to respect outcome-derived action
preferences while retaining compatible guidance from the solver's internal
branching heuristic.

\section{Experiments}
\label{sec:experiments}

In this section, we present a solver-level experimental analysis of CDCL branching to evaluate the effectiveness of \ours{}.


\vspace{-5px}
\paragraph{Datasets.}
We use three SAT families from G4SATBench~\citep{g4satbench}. These families are SR, Community Attachment (CA), and Popularity-Similarity (PS).
For each family, 16{,}000/2{,}000/2{,}000 instances are used for model training, validation, and testing, respectively.
Outcome-derived branching supervision is collected with an instrumented CaDiCaL
solver~\citep{BiereFallerFazekasFleuryFroleyks-CAV24} through independent
CDCL rollouts.
Dataset and supervision statistics are provided in
Appendix~\ref{app:datasets}, with solver sampling and evaluation settings in
Appendix~\ref{app:experimental_setup}.


\vspace{-5px}
\paragraph{Baselines.}
We compare \ours{} with native CaDiCaL (3.0.0),
NeuroCore~\citep{neurocore_SelsamB19},
GQSAT~\citep{nips/KurinGWC20},
SATformer~\citep{satformer_iccad23},
ImitSAT~\citep{zhang2026boolean}, and
RLAF~\citep{rlaf/abs-2505-16053}.
For a consistent solver-level comparison, all learning-based methods are
integrated with the same CaDiCaL solver while retaining their original
guidance mechanisms.
NeuroCore, GQSAT, and ImitSAT guide branching decisions during search,
whereas SATformer and RLAF provide one-shot guidance before search.

\vspace{-5px}
\paragraph{Metrics.}
We measure solver performance using the relative changes in conflicts
($\Delta_C$), decisions ($\Delta_D$), and propagations ($\Delta_P$) with
respect to native CaDiCaL.
Following ImitSAT~\citep{zhang2026boolean}, we also report median relative
propagation per problem (MRPP) and the one-percent win rate (W(1\%)).
Detailed metric definitions are provided in
Appendix~\ref{app:evaluation_metrics}.

\vspace{-5px}
\paragraph{Implementation Details.}
Our models use 10 message-passing layers with a hidden dimension of 128.
They are trained with AdamW~\citep{DBLP:conf/iclr/LoshchilovH19} for 50 epochs using a learning rate of $10^{-4}$.
We select $\lambda_{\mathrm{heur}}$ from $[0,0.2]$, where   
$\lambda_{\mathrm{heur}}=0$ removes non-dominated heuristic action supervision
and corresponds to \oursbase{}.
For the experiments on SR, CA, and PS, the 3-call and 5-call settings allow at
most three and five model requests per solver run, respectively, and each run
has a 100-second time limit.
Complete hyperparameter configurations and solver evaluation settings are
provided in Appendix~\ref{app:experimental_setup}.

\begin{table*}[!t]
\vspace{-10px}
\centering
\footnotesize
\caption{
Solver-level comparison on SR, CA, and PS.
Lower is better for $\Delta_C$, $\Delta_D$, $\Delta_P$, and MRPP,
while higher is better for W(1\%).
}
\label{tab:main_comparison}

\begin{tabular}{clrrrrr}
\toprule
\textbf{Dataset} & \multicolumn{1}{c}{\textbf{Methods}} &
$\Delta_C \downarrow$ & $\Delta_D \downarrow$ & $\Delta_P \downarrow$ &
MRPP $\downarrow$ & W(1\%) $\uparrow$ \\
\midrule

\multirow{11}{*}{SR}
& CaDiCaL~\citeyearpar{BiereFallerFazekasFleuryFroleyks-CAV24}
& 0.00\% & 0.00\% & 0.00\% & 1.000 & 0.00\% \\

& NeuroCore (3-call)~\citeyearpar{neurocore_SelsamB19}
& -32.61\% & -29.99\% & -33.38\% & 0.663 & 79.90\% \\

& NeuroCore (5-call)~\citeyearpar{neurocore_SelsamB19}
& -38.44\% & -35.29\% & -39.34\% & 0.602 & 84.30\% \\

& GQSAT (3-call)~\citeyearpar{nips/KurinGWC20}
& -20.74\% & -18.40\% & -20.64\% & 0.800 & 68.90\% \\

& GQSAT (5-call)~\citeyearpar{nips/KurinGWC20}
& -24.36\% & -21.80\% & -24.03\% & 0.774 & 70.80\% \\

& SATformer~\citeyearpar{satformer_iccad23}
& -12.22\% & -11.52\% & -13.02\% & 1.000 & 24.40\% \\

& ImitSAT (3-call)~\citeyearpar{zhang2026boolean}
& -0.30\% & -0.27\% & -0.37\% & 0.998 & 46.80\% \\

& ImitSAT (5-call)~\citeyearpar{zhang2026boolean}
& -0.35\% & -0.68\% & -0.35\% & 0.995 & 48.50\% \\

& RLAF~\citeyearpar{rlaf/abs-2505-16053}
& -26.99\% & -26.31\% & -27.13\% & 0.803 & 73.20\% \\

\cmidrule{2-7}

& \pancell{\ours{} (3-call)}
& \pancell{-38.01\%}
& \pancell{-35.17\%}
& \pancell{-37.49\%}
& \pancell{0.582}
& \pancell{85.70\%} \\

& \pancell{\ours{} (5-call)}
& \pancell{\textbf{-42.46\%}}
& \pancell{\textbf{-39.63\%}}
& \pancell{\textbf{-41.60\%}}
& \pancell{\textbf{0.534}}
& \pancell{\textbf{88.30\%}} \\

\midrule

\multirow{11}{*}{CA}
& CaDiCaL~\citeyearpar{BiereFallerFazekasFleuryFroleyks-CAV24}
& 0.00\% & 0.00\% & 0.00\% & 1.000 & 0.00\% \\

& NeuroCore (3-call)~\citeyearpar{neurocore_SelsamB19}
& -1.44\% & 2.51\% & 0.43\% & 1.029 & 30.40\% \\

& NeuroCore (5-call)~\citeyearpar{neurocore_SelsamB19}
& -6.12\% & 0.32\% & -5.29\% & 1.029 & 34.90\% \\

& GQSAT (3-call)~\citeyearpar{nips/KurinGWC20}
& -5.94\% & -2.08\% & -4.93\% & 1.006 & 41.50\% \\

& GQSAT (5-call)~\citeyearpar{nips/KurinGWC20}
& -6.68\% & -0.72\% & -4.45\% & 1.005 & 44.50\% \\

& SATformer~\citeyearpar{satformer_iccad23}
& -10.64\% & -4.30\% & -9.16\% & 1.000 & 35.60\% \\

& ImitSAT (3-call)~\citeyearpar{zhang2026boolean}
& 1.04\% & 2.25\% & 1.47\% & 1.011 & 22.30\% \\

& ImitSAT (5-call)~\citeyearpar{zhang2026boolean}
& 1.50\% & 3.33\% & 2.83\% & 1.015 & 24.30\% \\

& RLAF~\citeyearpar{rlaf/abs-2505-16053}
& \textbf{-23.42\%} & -14.04\% & \textbf{-21.39\%} & 0.909 & 63.50\% \\

\cmidrule{2-7}

& \pancell{\ours{} (3-call)}
& \pancell{-11.05\%}
& \pancell{-6.54\%}
& \pancell{-9.42\%}
& \pancell{0.961}
& \pancell{54.50\%} \\

& \pancell{\ours{} (5-call)}
& \pancell{-20.85\%}
& \pancell{\textbf{-14.48\%}}
& \pancell{-18.63\%}
& \pancell{\textbf{0.817}}
& \pancell{\textbf{70.20\%}} \\

\midrule

\multirow{11}{*}{PS}
& CaDiCaL~\citeyearpar{BiereFallerFazekasFleuryFroleyks-CAV24}
& 0.00\% & 0.00\% & 0.00\% & 1.000 & 0.00\% \\

& NeuroCore (3-call)~\citeyearpar{neurocore_SelsamB19}
& -17.42\% & -17.05\% & -16.32\% & 0.636 & 76.80\% \\

& NeuroCore (5-call)~\citeyearpar{neurocore_SelsamB19}
& -22.33\% & -21.72\% & -20.78\% & 0.585 & 77.70\% \\

& GQSAT (3-call)~\citeyearpar{nips/KurinGWC20}
& -2.18\% & -2.58\% & -2.86\% & 0.957 & 52.10\% \\

& GQSAT (5-call)~\citeyearpar{nips/KurinGWC20}
& -3.90\% & -3.91\% & -3.42\% & 0.962 & 51.30\% \\

& SATformer~\citeyearpar{satformer_iccad23}
& -15.41\% & -14.94\% & -13.02\% & 0.999 & 46.80\% \\

& ImitSAT (3-call)~\citeyearpar{zhang2026boolean}
& -1.26\% & -1.34\% & -1.72\% & 1.008 & 45.30\% \\

& ImitSAT (5-call)~\citeyearpar{zhang2026boolean}
& -1.75\% & -1.88\% & -2.21\% & 1.012 & 45.30\% \\

& RLAF~\citeyearpar{rlaf/abs-2505-16053}
& -20.12\% & -19.93\% & -18.55\% & 0.813 & 67.10\% \\

\cmidrule{2-7}

& \pancell{\ours{} (3-call)}
& \pancell{-33.23\%}
& \pancell{-32.17\%}
& \pancell{-31.34\%}
& \pancell{0.484}
& \pancell{86.00\%} \\

& \pancell{\ours{} (5-call)}
& \pancell{\textbf{-43.66\%}}
& \pancell{\textbf{-42.55\%}}
& \pancell{\textbf{-42.40\%}}
& \pancell{\textbf{0.417}}
& \pancell{\textbf{88.90\%}} \\

\bottomrule
\end{tabular}

\vspace{-15px}

\end{table*}


\paragraph{Exp-1. Comparison with Existing Methods for CDCL Branching.}
\label{par:exp_main}

Table~\ref{tab:main_comparison} compares \ours{} with native CaDiCaL and learning-based SAT branching baselines.
On SR and PS, \ours{} (5-call) improves the three aggregate search-cost metrics
over the best competing baseline by 2.26--4.34 and 20.83--21.62 percentage
points, respectively.
On CA, RLAF obtains lower aggregate conflicts and propagations, whereas
\ours{} (5-call) achieves better decisions, MRPP, and W(1\%).
For \ours{}, the 5-call setting generally improves over the 3-call setting;
we analyze the complete budget--performance relationship next.

The performance differences are consistent with the supervision used by each
method.
Structural and trajectory-level supervision can improve branching, but
provides less direct evidence for comparing actions at a dynamic solver state.
ImitSAT encodes the CNF and execution trace as sequences, and the difficulty
of modeling long sequences may help explain its weaker performance.
RLAF benefits from complete-run feedback and persistent global guidance, which
appears particularly effective on CA, although its weaker MRPP and W(1\%)
results suggest that these gains are less broadly distributed across instances.
In contrast, \ours{} learns from state-matched action outcomes and applies its
scores at dynamic decision points, yielding strong performance across the
three datasets.
Its smaller gains on CA are consistent with the lower coverage of preference
pairs reported in Table~\ref{tab:datasets}.
Additional analyses of SAT/UNSAT instances and neural inference cost are
provided in Appendices~\ref{app:sat_unsat} and~\ref{app:runtime}, respectively.

\paragraph{Exp-2. Effect of Neural Intervention Budget.}
\label{par:exp_intervention_budget}

\begin{figure}[t]
    \centering
    \includegraphics[width=0.95\textwidth]{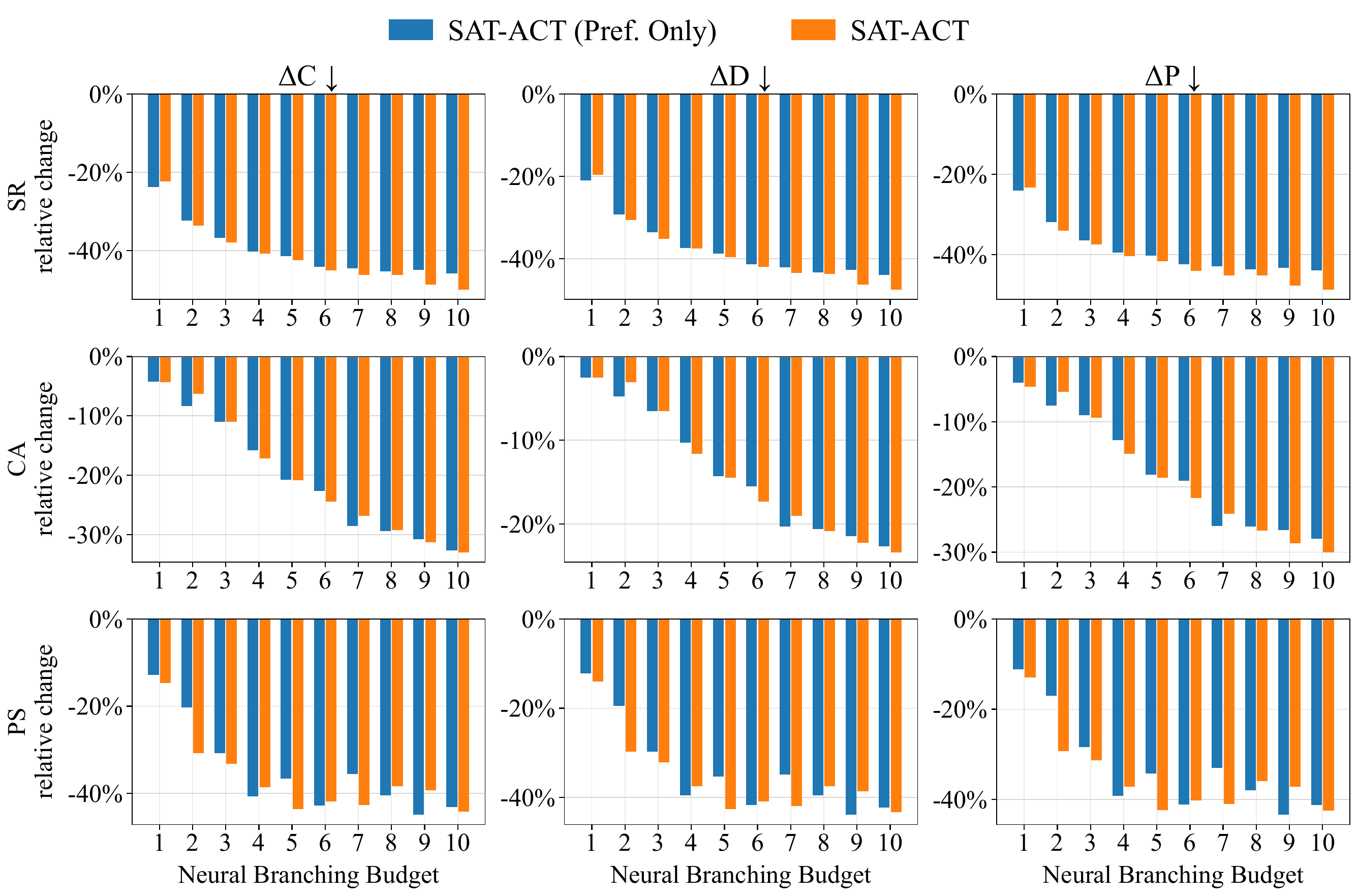}
    \caption{
    Effect of the neural intervention budget on SR, CA, and PS.
    Bars report relative changes in conflicts ($\Delta_C$), decisions
    ($\Delta_D$), and propagations ($\Delta_P$) with respect to native
    CaDiCaL.
    }
    \label{fig:intervention_budget}
\end{figure}

We next examine how solver performance changes with the maximum number of
neural model requests allowed during each solver run.
While Table~\ref{tab:main_comparison} reports the representative 3-call and
5-call settings, Figure~\ref{fig:intervention_budget} evaluates \oursbase{}
and \ours{} over the complete intervention-budget range.
Increasing the intervention budget generally leads to larger reductions in
conflicts, decisions, and propagations.
The effect is most pronounced on SR and CA, where all three search-cost
metrics improve substantially as the budget increases.
PS exhibits larger fluctuations across neighboring budgets, but higher
budgets still outperform very small budgets overall.
Across the tested range, \ours{} generally matches or outperforms
\oursbase{}, especially on SR and CA.
These results show that increasing the neural intervention budget can provide
additional reductions in search cost.

\begin{table}[t]
\vspace{-10px}
\centering
\footnotesize
\caption{
Incremental analysis of outcome comparison criteria.
Lower is better for all metrics.
}
\label{tab:preference_ablation}
\renewcommand{\arraystretch}{1.05}
\setlength{\tabcolsep}{2.0pt}
\begin{tabular}{lrrrrrrrrr}
    \toprule
    \textbf{Criteria}
    & \multicolumn{3}{c}{\textbf{SR}}
    & \multicolumn{3}{c}{\textbf{CA}}
    & \multicolumn{3}{c}{\textbf{PS}} \\
    \cmidrule(lr){2-4}
    \cmidrule(lr){5-7}
    \cmidrule(lr){8-10}
    & $\Delta_C \downarrow$
    & $\Delta_D \downarrow$
    & $\Delta_P \downarrow$
    & $\Delta_C \downarrow$
    & $\Delta_D \downarrow$
    & $\Delta_P \downarrow$
    & $\Delta_C \downarrow$
    & $\Delta_D \downarrow$
    & $\Delta_P \downarrow$ \\
    \midrule
    C1
    & -30.89\% & -28.33\% & -31.87\%
    & -9.67\% & -4.43\% & -8.34\%
    & -30.99\% & -29.76\% & -28.99\% \\
    C1 + C2
    & -39.23\% & -36.85\% & -38.39\%
    & -13.21\% & -8.05\% & -11.06\%
    & -34.32\% & -33.30\% & -31.97\% \\
    C1 + C2 + C3
    & \textbf{-41.49\%} & \textbf{-38.80\%} & \textbf{-40.30\%}
    & \textbf{-20.77\%} & \textbf{-14.30\%} & \textbf{-18.12\%}
    & \textbf{-36.55\%} & \textbf{-35.36\%} & \textbf{-34.24\%} \\
    \bottomrule
\end{tabular}
\vspace{-12px}
\end{table}

\vspace{-5px}
\paragraph{Exp-3. Ablation of Outcome Comparison Criteria.}
\label{par:exp_preference_ablation}

We analyze the contribution of the three outcome comparison criteria by progressively incorporating different forms of solving evidence.
We compare C1, C1+C2, and C1+C2+C3 under the 5-call setting while keeping the model, training procedure, and solver evaluation protocol unchanged.
All variants use preference-guided action ranking without heuristic action supervision.
The results are reported in Table~\ref{tab:preference_ablation}.
These results show that the complete C1+C2+C3 combination achieves the strongest reductions in conflicts, decisions, and propagations on all three datasets.
The contribution of each added criterion varies across datasets, but every addition improves all three search-cost metrics.
This pattern suggests that the three outcome comparison criteria provide complementary supervision, with terminal-closure information adding useful signals beyond completion and trajectory Pareto evidence.
A complete leave-one-out analysis is provided in Appendix~\ref{app:preference_ablation}.

\begin{table}[t]
\centering
\footnotesize
\caption{
Ablation of heuristic action supervision strategies.
Lower is better for all metrics.
}
\label{tab:pan_ablation}
\renewcommand{\arraystretch}{1.05}
\setlength{\tabcolsep}{2.2pt}
\begin{tabular}{lrrrrrrrrr}
    \toprule
    \textbf{Variant}
    & \multicolumn{3}{c}{\textbf{SR}}
    & \multicolumn{3}{c}{\textbf{CA}}
    & \multicolumn{3}{c}{\textbf{PS}} \\
    \cmidrule(lr){2-4}
    \cmidrule(lr){5-7}
    \cmidrule(lr){8-10}
    & $\Delta_C \downarrow$
    & $\Delta_D \downarrow$
    & $\Delta_P \downarrow$
    & $\Delta_C \downarrow$
    & $\Delta_D \downarrow$
    & $\Delta_P \downarrow$
    & $\Delta_C \downarrow$
    & $\Delta_D \downarrow$
    & $\Delta_P \downarrow$ \\
    \midrule
    Pref. Only
    & -41.49\% & -38.80\% & -40.30\%
    & -20.77\% & -14.30\% & -18.12\%
    & -36.55\% & -35.36\% & -34.24\% \\
    Heur. Only
    & -30.84\% & -27.80\% & -31.91\%
    & -9.42\% & -3.03\% & -5.43\%
    & -16.30\% & -15.59\% & -13.41\% \\
    Pref. + Heur.
    & -40.98\% & -38.13\% & -39.95\%
    & -17.52\% & -11.37\% & -13.58\%
    & -33.57\% & -32.65\% & -30.88\% \\
    Pref. + RG
    & -42.32\% & -39.50\% & \textbf{-41.85\%}
    & -19.97\% & -13.83\% & -17.36\%
    & -36.97\% & -36.09\% & -35.36\% \\
    \ours{}
    & \textbf{-42.46\%} & \textbf{-39.63\%} & -41.60\%
    & \textbf{-20.85\%} & \textbf{-14.48\%} & \textbf{-18.63\%}
    & \textbf{-43.66\%} & \textbf{-42.55\%} & \textbf{-42.40\%} \\
    \bottomrule
\end{tabular}
\vspace{-2px}
\end{table}

\vspace{-5px}

\paragraph{Exp-4. Ablation of Non-Dominated Heuristic Action Supervision.}
\label{par:exp_pan_ablation}

\begin{figure}[t]
    \centering
    \includegraphics[width=\textwidth]{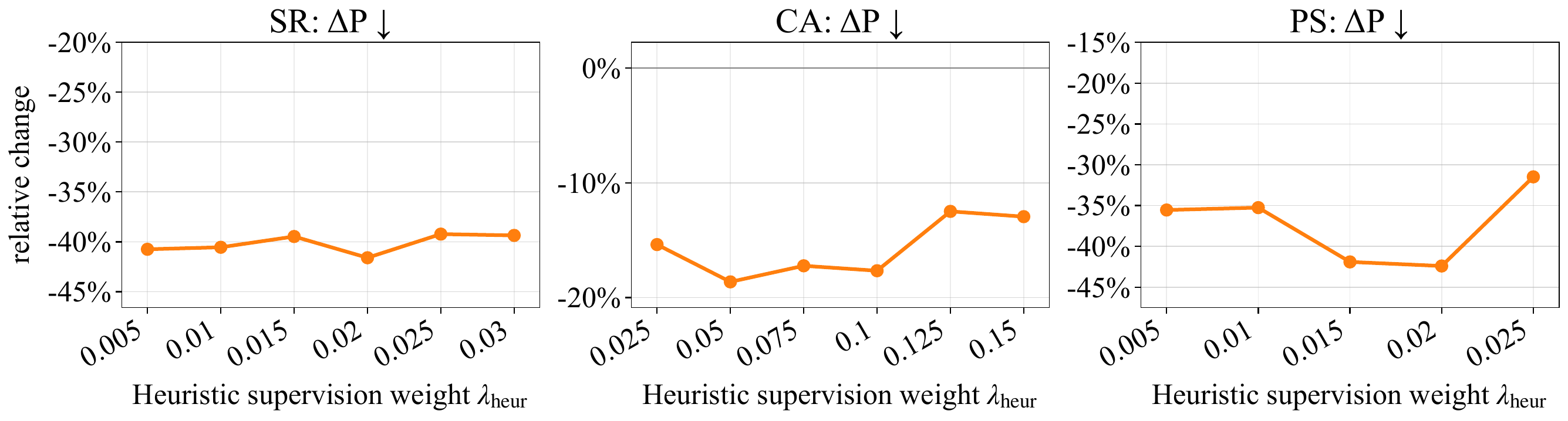}
    \caption{
    Sensitivity to the heuristic action supervision weight on SR, CA, and PS
    under the 5-call setting.
    The panels report relative changes in propagations ($\Delta_P$) with
    respect to native CaDiCaL.
    }
    \label{fig:pan_weight_main}
    \vspace{-5px}
\end{figure}

Under the 5-call setting, we compare \ours{} with four supervision variants.
\emph{Pref. Only} trains solely from outcome-derived action preferences and
removes heuristic action supervision.
\emph{Heur. Only} removes preference supervision and trains only on actions
selected by the solver's internal branching heuristic.
\emph{Pref. + Heur.} combines preference supervision with every heuristic
target, regardless of its relation to the evaluated actions.
\emph{Pref. + RG} instead retains heuristic targets through random gating.
The full \ours{} model uses outcome-derived preferences to retain a heuristic
target only when its action remains non-dominated.
Table~\ref{tab:pan_ablation} reports the results.
\emph{Heur. Only} and \emph{Pref. + Heur.} underperform
\emph{Pref. Only} across all three datasets, while random gating yields
inconsistent improvements.
\ours{} achieves the best $\Delta_C$ and $\Delta_D$ on all datasets and the
best $\Delta_P$ on CA and PS, while \emph{Pref. + RG} is slightly better on
SR.
These results support filtering heuristic targets with outcome-derived
preferences.
We further examine the sensitivity of non-dominated heuristic action
supervision to its weight under the 5-call setting.
Figure~\ref{fig:pan_weight_main} reports the relative change in propagations.
Performance on SR remains stable across the evaluated weights.
CA achieves its largest reductions at moderate weights around
$\lambda_{\mathrm{heur}}\in[0.05,0.1]$, while PS performs best around
$\lambda_{\mathrm{heur}}\in[0.015,0.02]$.
Overall, non-dominated heuristic action supervision remains effective over
a range of moderate weights.
Results for all three search-cost metrics are provided in
Appendix~\ref{app:pan_weight_full}.

\vspace{-5px}

\begin{table*}[t]
\vspace{-10px}
\centering
\footnotesize
\caption{
Solver integration results on SAT Competition 2023--2025 instances.
Solved is the sum of SAT and UNSAT results, and lower PAR-2 is better.
}
\label{tab:sat_solver_combination}
\renewcommand{\arraystretch}{1.05}
\setlength{\tabcolsep}{3.5pt}
\begin{tabular}{llrrrrr}
\toprule
\textbf{Dataset} & \textbf{Solver} &
\textbf{\#Solved} & \textbf{\#SAT} & \textbf{\#UNSAT} &
\textbf{\#Timeout} & \textbf{PAR-2 (s)} \\
\midrule
\multirow{2}{*}{SC-2023}
& CaDiCaL (Native)
& \textbf{251} & \textbf{116} & 135 & \textbf{149} & 4171.07 \\
& CaDiCaL (\ours{})
& \textbf{251} & 115 & \textbf{136} & \textbf{149} & \textbf{4150.55} \\
\midrule
\multirow{2}{*}{SC-2024}
& CaDiCaL (Native)
& 264 & \textbf{137} & 127 & 136 & 3989.14 \\
& CaDiCaL (\ours{})
& \textbf{266} & 135 & \textbf{131} & \textbf{134} & \textbf{3952.30} \\
\midrule
\multirow{2}{*}{SC-2025}
& CaDiCaL (Native)
& 253 & \textbf{144} & 109 & 147 & 4214.04 \\
& CaDiCaL (\ours{})
& \textbf{256} & 143 & \textbf{113} & \textbf{144} &
\textbf{4094.72} \\
\bottomrule
\end{tabular}
\vspace{-10px}
\end{table*}

\vspace{-5px}
\paragraph{Exp-5. Evaluation on SAT Competition Benchmarks.}
\label{par:exp_solver_combination}

We evaluate \ours{} with CaDiCaL on SAT Competition Main Track~\footnote{\url{https://satcompetition.github.io/}} instances in 2023--2025. Each year's competition contains 400 instances.
Both solvers use the same CaDiCaL configuration and per-instance time limit, differing only in neural branching guidance.
As shown in Table~\ref{tab:sat_solver_combination}, the guided solver matches
the number of solved instances on SC-2023 and solves two and three additional
instances on SC-2024 and SC-2025, respectively.
It also reduces PAR-2 on all three benchmark sets.
The independent training setup, evaluation protocol, and neural-call overhead measurements are provided in
Appendix~\ref{app:sat_solver_combination}.

\vspace{-5px}

\section{Related Work}
\label{sec-related}

In this section, we review related studies on learning-based approaches for SAT solving.

\textbf{End-to-End Learning for SAT Solving.}
End-to-end approaches learn to reason directly 
over neural representations of SAT instances.
Early methods focus on CNF representations.
\citet{corr/BunzL17} apply GNNs to satisfiability prediction using a
signed-literal graph with clause co-occurrence and complementarity edges.
NeuroSAT~\citep{neurosat_SelsamLBLMD19} uses a literal--clause graph
with connections between complementary literals and performs iterative
message passing between literal and clause nodes.
Permutation-invariant architectures have also been developed to learn
directly from CNF formulas~\citep{aaai/CameronCHL20}.
Another group of methods targets circuit representations.
DG-DAGRNN~\citep{iclr/AmizadehMW19} represents circuit SAT instances as
directed acyclic graphs and learns satisfying assignments with gated recursive
networks.
DeepSAT~\citep{dac/LiSLKCX23} instead operates on AND-inverter graphs and
incorporates polarity-aware representations for Boolean reasoning.
Neural reasoning has also been extended to related satisfiability tasks.
HyperSAT~\citep{hypersat25} develops an unsupervised hypergraph neural
framework for Weighted MaxSAT, while NSNet~\citep{nips/LiS22} formulates SAT
and \#SAT as probabilistic inference with neural belief propagation.
These approaches perform prediction, assignment construction, or
optimization directly from the input instance.
Our work instead retains the CDCL solving process and learns to guide
branching decisions from dynamic solver states.

\textbf{Learning-Guided SAT Solving.}
Learning-guided methods integrate neural predictions into established SAT
solvers.
Beyond branching, learned models have been used to guide restart
policies~\citep{sat/LiangOMTLG18} and clause deletion~\citep{dac/LiuXPYZYH024}.
Learning-guided techniques have also been applied to stochastic local search
for variable selection~\citep{nips/YolcuP19} and assignment
initialization~\citep{ijcai/ZhangSZLCXZ20}.
GraSS~\citep{kdd/ZhangCCGXZ0HCY24} learns solver selection from graph-based
representations. 
For CDCL branching, several methods learn structural properties of SAT
formulas.
NeuroCore~\citep{neurocore_SelsamB19} predicts unsat-core variables to guide
variable selection, while SATformer~\citep{satformer_iccad23} improves this
prediction through transformer-based representations.
\citet{sun2026unsat} develop polarity-aware representations over
hypergraphs for this task, and
NeuroBack~\citep{NeuroBack_iclr24} predicts backbone phases for phase
initialization.
Other methods obtain branching guidance from the solving process.
Graph-Q-SAT~\citep{nips/KurinGWC20} learns through reinforcement
learning and solver interaction.
RLAF~\citep{rlaf/abs-2505-16053} uses solver-level algorithm feedback to
predict variable weights and polarities, while
ImitSAT~\citep{zhang2026boolean} learns branching decisions from CDCL
execution traces.

Candidate evaluation under a shared search context also has precedents in
combinatorial search.
Classical SAT lookahead solvers probe candidate branches through
propagation-based effects~\citep{biere2021look}, while strong
branching in MILP evaluates candidate variables under a
shared branch-and-bound context~\citep{khalil2016learning,gasse2019exact}.
These precedents demonstrate the value of comparing candidate decisions under
a shared search context.
Inspired by these precedents, we construct partial action preferences from
state-matched rollout outcomes at dynamic CDCL states and use them to train a
branching policy that operates without online candidate probing.

\vspace{-5px}
\section{Conclusion}
\label{sec-conclusion}
\vspace{-5px}

In this paper, we presented \ours{}, an outcome-derived learning framework for
CDCL branching.
The framework compares independent rollouts initiated from the same solver
state and applies three complementary outcome comparison criteria to derive
partial preferences without a predefined scalar action-quality score.
A state-conditioned branching action network learns branching priorities from
these preferences, while non-dominated heuristic action supervision retains
solver-heuristic guidance only when the selected choice remains non-dominated.

Experiments on three SAT families show that \ours{} consistently reduces CDCL
search costs relative to native CaDiCaL and compares favorably with existing
learning-based methods.
It also improves PAR-2 across three SAT Competition benchmarks and increases
overall solving coverage.
Ablation studies further demonstrate the contributions of the outcome
comparison criteria and non-dominated heuristic action supervision.
Future work includes reducing offline rollout costs, improving candidate and
intervention selection, and extending the framework to broader SAT
distributions and solver configurations.






\bibliography{ref}
\bibliographystyle{iclr2027_conference}

\appendix
\newpage

\section{Additional Details of Outcome Collection and Action Preference Extraction}
\label{app:outcome_preference_details}

This section provides additional details of outcome collection and action preference extraction in Section~\ref{subsec:outcome_preferences}.
We describe how decision states and candidate branching actions are selected, how independent rollouts are run from the same solver state, how the rollout budget is defined, and how terminal-closure levels are determined.

\subsection{Decision-State Sampling}

For each solver run, we retain a small number of decision states distributed across different stages of the search.
Early decision states are always represented, while additional states are sampled from later parts of the search trajectory.
We also retain states reached after restarts so that the collected states cover both the initial search phase and later search contexts.

Each retained state corresponds to the solver state $S_t$ immediately before a branching decision.
The state contains the active clauses, current partial assignment, decision trail, variable activities, saved phases, and other dynamic search information accumulated during CDCL solving.

\subsection{Candidate Branching Action Sampling and Rollouts}
\label{app:candidate_evaluation}

Let $\mathcal{B}_t$ denote the legal branching actions at $S_t$.
We construct a bounded candidate set
\begin{equation}
    \mathcal{A}_t
    =
    \{a_t^1,\ldots,a_t^{K_t}\},
    \qquad
    \mathcal{A}_t\subseteq\mathcal{B}_t .
\end{equation}

\paragraph{Candidate construction.}
We combine the branching variable selected by the solver's internal heuristic with activity-based,
residual Jeroslow--Wang (JW)~\citep{DBLP:journals/amai/JeroslowW90}, and pseudorandom proposals.
The heuristic variable preserves the solver's reference action.
Activity reflects accumulated search history, residual JW captures the current unsatisfied-clause structure, and random proposals broaden coverage beyond the heuristic rankings.
This mixture is a practical design under a limited rollout budget;
we do not claim that these sources are optimal or exhaustive.

We first retain the heuristic variable.
We then construct three proposal pools with up to eight legal branching variables in each pool.
The pools contain the highest-activity variables, the highest-scoring variables under residual JW, and a reproducible pseudorandom selection.
For residual JW, each unsatisfied active clause contributes
$2^{-r}$ to each of its unassigned active literals, where $r$ is
the number of such literals in the clause.
A variable is ranked by the larger of its two literal scores.

Starting with the heuristic variable, we add previously unselected
variables in round-robin order across the activity, JW, and random proposals until
eight distinct variables have been retained or the pools are exhausted.
This interleaving prevents an earlier proposal source from filling
the entire candidate budget before the others are considered.
We run a separate rollout for each polarity of every retained variable,
yielding $K_t\leq16$ distinct branching actions.
The heuristic action is therefore retained in $\mathcal{A}_t$.
An additional unforced solver continuation is recorded separately
as a baseline and does not define another action in $\mathcal{A}_t$.

\paragraph{State reconstruction.}
For every action rollout, we reconstruct the target
pre-action state by deterministic re-execution rather than by
loading a solver snapshot.
We launch a fresh solver process, load the original CNF, and
re-execute the original search using the same solver binary,
configuration, and random seed as in state collection.
No candidate intervention is applied before reaching the target.
Re-executing the search prefix regenerates the active clause
database, partial assignment, decision trail, variable activities,
and saved phases through the solver's normal updates; it is not
merely a reassignment of recorded decision literals.
At the target decision point, we check consistency with the recorded
state locator before applying the candidate action.
Variable activities and saved phases are reconstructed through
deterministic execution rather than separately loaded or directly
checked by the locator.
For offline data collection, we use CaDiCaL 3.0.0 in
\texttt{plain} mode with seed 1 and \texttt{stabilizeonly=1}.

\paragraph{Action intervention and rollout.}
At the reconstructed state $S_t$, we override exactly one branching
decision with the candidate literal $a\in\mathcal{A}_t$.
This is an ordinary, backtrackable decision, not an assumption or
an added unit clause.
Subsequent decisions follow the solver's internal CDCL procedure, and normal
CDCL backtracking may undo the intervention and earlier non-root
assignments.
Thus, a completed rollout reports the global SAT/UNSAT result
of the original instance, not the satisfiability of a formula
conditioned on the candidate literal.
Completed rollouts for the same instance therefore share its
global SAT/UNSAT result; actions differ in whether they finish
within the budget and in the search effort required.

Each rollout records its solving status and the numbers of
conflicts, propagations, decisions, and restarts, corresponding to
$O_t(a)$ in Section~\ref{subsec:outcome_preferences}.
These counters are measured relative to the pre-action state,
excluding the replay prefix, and the decision count includes the
intervened decision itself.
A rollout is labeled SAT or UNSAT if it reaches a terminal
result within the rollout budget, and incomplete if a budget
limit terminates it before a terminal result is reached.
All candidate actions rolled out from the same state use the same
solver configuration and rollout budget, as specified in
Appendix~\ref{app:rollout_budget}.
Preference pairs are extracted only between actions rolled out
from the same pre-action state; no cross-state or cross-instance
cost comparisons are used to assign preference labels.

\subsection{Rollout Budget}
\label{app:rollout_budget}

To bound the cost of action rollouts, each rollout is executed under a fixed rollout budget.
We represent the budget as
\begin{equation}
    \mathbf{b}_{\mathrm{roll}}
    =
    \left(
        C_{\mathrm{lim}},
        D_{\mathrm{lim}},
        P_{\mathrm{lim}},
        T_{\mathrm{lim}}
    \right),
\end{equation}
where $C_{\mathrm{lim}}$, $D_{\mathrm{lim}}$, and $P_{\mathrm{lim}}$
denote the maximum numbers of conflicts, decisions, and propagations allowed during the rollout, respectively,
and $T_{\mathrm{lim}}$ denotes the maximum rollout time.
For offline supervision collection, we use propagation horizons of 1,000, 15,000, and 5,000 for CA, PS, and SR, respectively, with a 300-second rollout timeout and no conflict or decision limit.

These quantities are measured from the point at which the candidate branching action is applied.
The rollout terminates when the solver reaches SAT or UNSAT or when one of the prescribed limits is reached.
Accordingly, the solving status is defined as
\begin{equation}
y_t(a)
=
\begin{cases}
\mathrm{SAT},
& \text{if the rollout reaches a satisfying assignment},\\
\mathrm{UNSAT},
& \text{if the rollout proves unsatisfiability},\\
\mathrm{incomplete},
    & \text{if a rollout limit is reached before termination}.
\end{cases}
\end{equation}

The same rollout budget is applied to the candidate actions rolled out from the same solver state.
For an incomplete rollout, the solver statistics observed before reaching the rollout limit are retained as part of $O_t(a)$.

\subsection{Terminal Closure}
\label{app:terminal_closure}

Terminal closure provides additional evidence about how directly a completed rollout reaches its final SAT or UNSAT result.
The decision count $D_t(a)$ includes the forced branching decision itself.
Therefore,
\begin{equation}
    D_t(a)=1
\end{equation}
indicates that no additional branching decision is required after applying $a$.

For a SAT rollout, we define the terminal-closure level as
\begin{equation}
\kappa_t(a)
=
\begin{cases}
2,
& C_t(a)=0,\;R_t(a)=0,\;D_t(a)=1,\\
1,
& C_t(a)=0,\;R_t(a)=0,\;D_t(a)>1,\\
0,
& C_t(a)>0\;\text{or}\;R_t(a)>0.
\end{cases}
\end{equation}

The highest level corresponds to a satisfying assignment reached without any additional branching decision, conflict, or restart.
The intermediate level corresponds to a conflict-free and restart-free rollout that still requires additional branching.
A rollout containing a conflict or restart receives the lowest SAT closure level.

For an UNSAT rollout, we define
\begin{equation}
\kappa_t(a)
=
\begin{cases}
1,
& D_t(a)=1,\\
0,
& D_t(a)>1.
\end{cases}
\end{equation}

The higher UNSAT closure level is assigned when the forced action leads to the terminal contradiction without another branching decision.
Otherwise, the action receives the lower level.
Conflicts are not used to distinguish UNSAT closure levels because reaching an UNSAT result in CDCL necessarily involves conflict-based reasoning.

For two completed actions $a_i$ and $a_j$, a higher terminal-closure level provides preference evidence in favor of that action.
Formally,
\begin{equation}
    \kappa_t(a_i)>\kappa_t(a_j)
    \quad\Longrightarrow\quad
    a_i\succ_{S_t}a_j .
\end{equation}
Equal closure levels provide no additional preference evidence.

\section{Node Features for Dynamic CDCL States}
\label{app:state_features}

We provide the detailed node features used for the dynamic CDCL state representation in Section~\ref{subsec:learning_cdcl_branching}.
The features describe the current assignment, local formula structure, and global search context.
Normalized quantities are denoted by $\widetilde{\cdot}$.

\subsection{Literal Features}

For a literal $\ell$ associated with variable $v$, we use
\begin{equation}
    \mathbf{x}^{L}_{\ell}
    =
    \left[
        u_{\ell},
        t_{\ell},
        f_{\ell},
        \widetilde{d}^{\,\mathrm{asg}}_v,
        \widetilde{o}_{\ell},
        \log(1+o_{\ell}),
        \log(1+|\mathcal{V}_t|),
        \log(1+|\mathcal{C}_t|),
        \rho_t,
        \widetilde{d}_t
    \right].
\end{equation}

Here, $u_{\ell}$, $t_{\ell}$, and $f_{\ell}$ indicate whether the literal is unassigned, true, or false under the current partial assignment.
$\widetilde{d}^{\,\mathrm{asg}}_v$ denotes the normalized assignment level of variable $v$.
$o_{\ell}$ and $\widetilde{o}_{\ell}$ denote the occurrence count of $\ell$ in the active clauses and its normalized counterpart.
$\rho_t$ is the fraction of currently assigned variables, and $\widetilde{d}_t$ is the normalized current decision level.
The graph-size terms provide the current numbers of variables and active clauses on a logarithmic scale.

\subsection{Clause Features}

For a clause $c$, let $n_c^{T}$, $n_c^{F}$, and $n_c^{U}$ denote the numbers of true, false, and unassigned literals.
Its feature vector is
\begin{equation}
    \mathbf{x}^{C}_{c}
    =
    \left[
        \log(1+|c|),
        \frac{n_c^{T}}{|c|},
        \frac{n_c^{F}}{|c|},
        \frac{n_c^{U}}{|c|},
        q_c^{\mathrm{sat}},
        q_c^{(1)},
        q_c^{(2)}
    \right].
\end{equation}

Here, $q_c^{\mathrm{sat}}$ indicates whether the clause is currently satisfied, while $q_c^{(1)}$ and $q_c^{(2)}$ indicate unsatisfied clauses with one and two unassigned literals, respectively.

\section{Details of Polarity-Aware Branching Action Scoring}
\label{app:action_scoring}

This section provides additional details of the representation modules used for branching-action scoring in Section~\ref{subsec:learning_cdcl_branching}.

For each variable $v$, polarity-symmetric variable encoding constructs
\begin{equation}
    \mathbf{u}_v
    =
    \phi_{\mathrm{sym}}
    \left(
        \left[
            \mathbf{h}_{v^{+}}+\mathbf{h}_{v^{-}}
            \,;\,
            \left|
                \mathbf{h}_{v^{+}}-\mathbf{h}_{v^{-}}
            \right|
        \right]
    \right).
\end{equation}

Graph-level context aggregation first pools the variable and clause representations
\begin{equation}
    \mathbf{m}_t
    =
    \frac{
        \sum_{v\in\mathcal{V}_t}\mathbf{u}_v
        +
        \sum_{c\in\mathcal{C}_t}\mathbf{h}_c
    }{
        |\mathcal{V}_t|+|\mathcal{C}_t|
    }.
\end{equation}

The pooled representation is combined with a learnable global token to obtain the graph-level context
\begin{equation}
    \mathbf{g}_t
    =
    \Phi_{\mathrm{graph}}
    \left(
        \mathbf{z}_g,
        \phi_{\mathrm{ctx}}(\mathbf{m}_t)
    \right),
\end{equation}
where $\mathbf{z}_g$ is a learnable graph token, $\phi_{\mathrm{ctx}}$ projects the pooled representation, and $\Phi_{\mathrm{graph}}$ is a learnable context aggregation module.

Graph-conditioned variable encoding then incorporates this context into each variable representation
\begin{equation}
    \widetilde{\mathbf{u}}_v
    =
    \Phi_{\mathrm{var}}
    \left(
        \mathbf{u}_v,
        \mathbf{g}_t
    \right),
\end{equation}
where $\Phi_{\mathrm{var}}$ is a shared learnable conditioning module.
In our implementation, both context aggregation modules use GRUs~\citep{cho-etal-2014-learning}.

Finally, for a branching action $a$ associated with variable $v(a)$, the action score is
\begin{equation}
    s_\theta(S_t,a)
    =
    \phi_{\mathrm{score}}
    \left(
        \left[
            \mathbf{h}_a
            \,;\,
            \widetilde{\mathbf{u}}_{v(a)}
            \,;\,
            \mathbf{g}_t
        \right]
    \right),
\end{equation}
where $\phi_{\mathrm{score}}$ is shared across both polarities.
The action-specific literal representation $\mathbf{h}_a$ distinguishes the two polarities of the same variable, while $\widetilde{\mathbf{u}}_{v(a)}$ and $\mathbf{g}_t$ provide variable-level and graph-level context.

\section{Datasets and Supervision Statistics}
\label{app:datasets}

We use the SR, Community Attachment (CA), and Popularity-Similarity (PS)
families from G4SATBench~\citep{g4satbench}.
Each family contains both SAT and UNSAT instances.
The instances are generated using the SR generator from
NeuroSAT~\citep{neurosat_SelsamLBLMD19}, the CA
model~\citep{ca_giraldez}, and the PS model~\citep{ps_giraldez}.
For each family, we use 16{,}000/2{,}000/2{,}000 instances
for training, validation, and offline testing, respectively,
with balanced SAT and UNSAT instances.
The supervision statistics below aggregate the training, validation, and
offline test splits.

For each instance, an instrumented CaDiCaL
solver~\citep{BiereFallerFazekasFleuryFroleyks-CAV24}
retains up to six decision states and runs rollouts for up to sixteen
candidate branching actions at each retained state.
The resulting upper bounds are 360{,}000 retained states and
5.76 million independent CDCL rollouts.
Recording one additional unforced solver continuation at each
state gives an upper bound of 6.12 million solver runs.
These are design upper bounds.
An instance may yield fewer
than six retained states, and a state may contain fewer than
sixteen candidate actions.
The unforced solver continuation is recorded separately and does not
define an additional branching action.

Each action rollout reconstructs the target state
by re-executing the original search prefix, as described in
Appendix~\ref{app:candidate_evaluation}.
The three outcome comparison criteria in
Section~\ref{subsec:outcome_preferences} extract partial action preferences
from the resulting outcomes.
Raw rollout generation was executed sequentially by
family with 20 workers, taking 2 h 22 min for CA, 2 h 07 min
for SR, and 2 h 27 min for PS, or approximately 6 h 55 min
in total.

Table~\ref{tab:datasets} summarizes the CNF statistics and
the resulting outcome-derived supervision.

\begin{table*}[!t]
\centering
\footnotesize
\caption{
Dataset and outcome-derived supervision statistics.
CNF statistics are reported per instance.
Decision states and rolled-out actions are reported after supervision filtering.
A state is included only if it produces at least one preference pair.
Rolled-out actions count candidate branching actions and exclude the additional unforced solver baseline.
Criterion coverage is measured over all action pairs within these retained states, and the three criteria are not mutually exclusive.
}
\label{tab:datasets}
\renewcommand{\arraystretch}{1.08}

\begin{tabular*}{\textwidth}{@{\extracolsep{\fill}}lrrr}
\toprule
\textbf{Statistic} & \textbf{SR} & \textbf{CA} & \textbf{PS} \\
\midrule

\multicolumn{4}{l}{\textit{Formula Statistics (Mean / Min / Max)}} \\
Variables
& 119.78 / 40 / 200
& 120.81 / 40 / 200
& 118.70 / 40 / 200 \\

Clauses
& 652.90 / 159 / 1{,}233
& 1{,}671.15 / 175 / 2{,}999
& 823.15 / 224 / 1{,}599 \\

Clause/Variable Ratio
& 5.48 / 3.54 / 8.72
& 13.64 / 4.38 / 15.00
& 6.93 / 5.13 / 8.00 \\

\midrule

\multicolumn{4}{l}{\textit{Outcome-Derived Supervision}} \\
Instances
& 20{,}000
& 20{,}000
& 20{,}000 \\

Decision States
& 107{,}448
& 109{,}609
& 108{,}751 \\

Rolled-Out Actions
& 1{,}712{,}508
& 1{,}753{,}744
& 1{,}738{,}400 \\

Candidate Pairs
& 12{,}817{,}702
& 13{,}153{,}080
& 13{,}031{,}332 \\

Preference Pairs
& 9{,}970{,}806
& 9{,}336{,}502
& 10{,}274{,}426 \\

\midrule

\multicolumn{4}{l}{\textit{Criterion Coverage}} \\
C1 Completion
& 521{,}391 (4.07\%)
& 560{,}030 (4.26\%)
& 311{,}065 (2.39\%) \\

C2 Trajectory Pareto
& 9{,}411{,}303 (73.42\%)
& 8{,}756{,}335 (66.57\%)
& 9{,}912{,}676 (76.07\%) \\

C3 Terminal Closure
& 270{,}985 (2.11\%)
& 105{,}060 (0.80\%)
& 268{,}018 (2.06\%) \\


Combined Coverage
& 77.79\%
& 70.98\%
& 78.84\% \\

\bottomrule
\end{tabular*}
\end{table*}

\section{Evaluation Metrics}
\label{app:evaluation_metrics}

We evaluate solver search cost using conflicts ($C$), decisions ($D$), and propagations ($P$).
For these quantities, we report the percentage changes
$\Delta_C$, $\Delta_D$, and $\Delta_P$ relative to native CaDiCaL, computed from the aggregated counts over the evaluated instances.
Negative values indicate reductions in solver cost.
Following ImitSAT~\citep{zhang2026boolean}, we additionally report the median relative propagation per problem (MRPP) and the one-percent win rate W(1\%).
Let $\mathcal{I}$ denote the set of evaluated instances, and let $P_i$ and $P_i^{\mathrm{nat}}$ denote the propagation counts of the evaluated method and native CaDiCaL on instance $i$, respectively.
MRPP is defined as
\begin{equation}
    \mathrm{MRPP}
    =
    \operatorname{median}_{i\in\mathcal{I}}
    \left(
        \frac{P_i}{P_i^{\mathrm{nat}}}
    \right).
\end{equation}
Values below one indicate fewer propagations than native CaDiCaL on the median instance.

To capture instance-wise improvements, we further report the one-percent win rate.
Following the definition in ImitSAT, with margin $\delta=0.01$,
\begin{equation}
    \mathrm{W}(1\%)
    =
    \frac{100}{|\mathcal{I}|}
    \sum_{\substack{i\in\mathcal{I}\\P_i^{\mathrm{nat}}>0}}
    \mathbb{I}
    \left[
        P_i
        \leq
        (1-\delta)P_i^{\mathrm{nat}}
    \right].
\end{equation}
W(1\%) therefore measures the percentage of evaluated instances for which the method reduces propagations by at least $1\%$ relative to native CaDiCaL.
Higher values are better.

\section{Experimental Setup}
\label{app:experimental_setup}

\subsection{Implementation Details}

Our models are implemented in PyTorch with 10 message-passing rounds and two-layer MLPs.
The hidden dimension is 128.
Models are trained using AdamW for 50 epochs with a learning rate of $10^{-4}$.
For SR, CA, and PS, we use a batch size of 128, select the weight decay from
$\{0,10^{-5},5\times10^{-5},10^{-4}\}$, and retain the checkpoint with the
lowest validation loss.
The weight $\lambda_{\mathrm{heur}}$ for non-dominated heuristic action supervision is selected from $[0,0.2]$.
Setting $\lambda_{\mathrm{heur}}=0$ removes this supervision and corresponds to \oursbase{}.
All hyperparameters are selected exclusively on the validation split and fixed before solver-level test evaluation.
All models are trained on a machine equipped with an Intel Xeon Gold 6148 CPU
at 2.40\,GHz and an NVIDIA Tesla V100 PCIe GPU with 32\,GB memory.
For the SR, CA, and PS experiments, CaDiCaL runs on the CPU, while neural
inference is performed on the GPU.
Code is available at
\url{https://anonymous.4open.science/r/SAT-ACT-Review-1DA2}.

\subsection{Solver Evaluation}

All solver-level evaluations use CaDiCaL 3.0.0 with
\texttt{stabilizeonly=1} and \texttt{preprocesslight=0}, while the remaining
solver options retain their default values.
For neural intervention, both the node and cell size checks require
$2|\mathcal{V}_t|+|\mathcal{C}_t|+|\mathcal{E}_t|\leq2{,}000{,}000$ for the
literal--clause representation of each active CNF.

For each SAT family, we randomly sample 200 instances from the test split, with equal numbers of SAT and UNSAT instances.
We repeat the sampling and solver evaluation five times, with all methods evaluated on the same sampled instances and under the same solver settings.
The reported results are averaged across the five runs.

Each solver run is limited to 100 seconds.
Native CaDiCaL without neural guidance serves as the reference solver for computing $\Delta_C$, $\Delta_D$, $\Delta_P$, MRPP, and W(1\%).

For online branching methods, the 3-call and 5-call settings allow at most three and five model requests per solver run, respectively.
Each request counts toward this intervention budget regardless of whether the proposed branching action is ultimately applied by the solver.

\section{Additional Experimental Results}

\subsection{Analysis of SAT and UNSAT Results}
\label{app:sat_unsat}

We separately evaluate \ours{} on SAT and UNSAT instances under both 3-call and 5-call settings.
We report the same solver-level cost metrics used in the main comparison.

Table~\ref{tab:sat_unsat} shows that \ours{} consistently reduces conflicts, decisions, and propagations on both SAT and UNSAT instances across all three datasets.
The magnitude of the improvement varies across datasets and termination regimes.
This may reflect the different search characteristics of SAT and UNSAT solving, since SAT solving can terminate after finding one satisfying assignment, whereas UNSAT solving requires completing a proof.
Overall, the results show that \ours{} remains effective under both termination regimes.

\begin{table*}[!t]
\centering
\footnotesize
\caption{
SAT and UNSAT subset analysis of \ours{}.
Lower is better for all metrics.
}
\label{tab:sat_unsat}

\setlength{\tabcolsep}{4pt}

\begin{tabular}{clrrrrrr}
\toprule
\textbf{Dataset}
& \multicolumn{1}{c}{\textbf{Budget}}
& \multicolumn{3}{c}{\textbf{SAT}}
& \multicolumn{3}{c}{\textbf{UNSAT}} \\
\cmidrule(lr){3-5}
\cmidrule(lr){6-8}
&
&
$\Delta_C \downarrow$
& $\Delta_D \downarrow$
& $\Delta_P \downarrow$
& $\Delta_C \downarrow$
& $\Delta_D \downarrow$
& $\Delta_P \downarrow$ \\
\midrule

\multirow{2}{*}{SR}
& 3-call
& -45.82\% & -36.35\% & -43.17\%
& -33.70\% & -34.11\% & -34.08\% \\

& 5-call
& -51.43\% & -41.42\% & -47.33\%
& -37.34\% & -37.95\% & -37.88\% \\

\midrule

\multirow{2}{*}{CA}
& 3-call
& -14.11\% & -7.23\% & -12.35\%
& -6.77\% & -5.02\% & -4.65\% \\

& 5-call
& -18.47\% & -9.46\% & -15.82\%
& -23.87\% & -25.94\% & -22.95\% \\

\midrule

\multirow{2}{*}{PS}
& 3-call
& -40.53\% & -37.10\% & -38.53\%
& -26.68\% & -27.14\% & -23.72\% \\

& 5-call
& -59.47\% & -54.92\% & -58.22\%
& -29.31\% & -29.98\% & -26.15\% \\

\bottomrule
\end{tabular}
\end{table*}

\begin{table*}[!t]
\centering
\footnotesize
\caption{
Leave-one-out analysis of outcome comparison criteria.
Lower is better for all metrics.
}
\label{tab:preference_ablation_full}

\begin{tabular}{clrrrrrr}
\toprule
\textbf{Dataset}
& \multicolumn{1}{c}{\textbf{Comparison Criteria}}
& \multicolumn{3}{c}{\textbf{3-call}}
& \multicolumn{3}{c}{\textbf{5-call}} \\
\cmidrule(lr){3-5}
\cmidrule(lr){6-8}
&
& $\Delta_C \downarrow$
& $\Delta_D \downarrow$
& $\Delta_P \downarrow$
& $\Delta_C \downarrow$
& $\Delta_D \downarrow$
& $\Delta_P \downarrow$ \\
\midrule

\multirow{5}{*}{SR}
& C1 + C2 + C3
& -36.84\% & -33.57\% & -36.45\%
& -41.49\% & -38.80\% & -40.30\% \\

& w/o C1
& -26.66\% & -24.56\% & -26.00\%
& -40.01\% & -37.55\% & -39.03\% \\

& w/o C2
& -29.71\% & -27.03\% & -30.00\%
& -32.76\% & -30.81\% & -32.35\% \\

& w/o C3
& -28.85\% & -27.23\% & -28.76\%
& -39.23\% & -36.85\% & -38.39\% \\

& C1 only
& -30.30\% & -27.60\% & -30.91\%
& -30.89\% & -28.33\% & -31.87\% \\

\midrule

\multirow{5}{*}{CA}
& C1 + C2 + C3
& -10.99\% & -6.51\% & -8.97\%
& -20.77\% & -14.30\% & -18.12\% \\

& w/o C1
& -7.78\% & -4.24\% & -6.48\%
& -14.74\% & -9.33\% & -12.51\% \\

& w/o C2
& -7.94\% & -3.64\% & -6.95\%
& -8.97\% & -3.17\% & -5.93\% \\

& w/o C3
& -9.83\% & -5.39\% & -7.55\%
& -13.21\% & -8.05\% & -11.06\% \\

& C1 only
& -5.83\% & -2.35\% & -5.32\%
& -9.67\% & -4.43\% & -8.34\% \\

\midrule

\multirow{5}{*}{PS}
& C1 + C2 + C3
& -30.75\% & -29.68\% & -28.38\%
& -36.55\% & -35.36\% & -34.24\% \\

& w/o C1
& -28.64\% & -27.60\% & -25.95\%
& -28.08\% & -27.35\% & -24.40\% \\

& w/o C2
& -19.51\% & -18.73\% & -16.39\%
& -31.74\% & -30.80\% & -30.14\% \\

& w/o C3
& -26.63\% & -25.87\% & -23.62\%
& -34.32\% & -33.30\% & -31.97\% \\

& C1 only
& -22.54\% & -21.52\% & -20.14\%
& -30.99\% & -29.76\% & -28.99\% \\

\bottomrule
\end{tabular}
\end{table*}

\subsection{Additional Ablations of Outcome Comparison Criteria}
\label{app:preference_ablation}

We further analyze the three outcome comparison criteria using leave-one-out ablations.
C1, C2, and C3 denote completion, trajectory Pareto, and terminal closure, respectively.
In addition to the full C1+C2+C3 combination, we remove each criterion individually and include the C1-only variant used in the incremental analysis.
All variants share the same model architecture and solver evaluation protocol,
with preference-guided action ranking as the sole training objective.
Table~\ref{tab:preference_ablation_full} provides a consistent picture across datasets and intervention budgets.
The full C1+C2+C3 combination achieves the largest reductions in conflicts, decisions, and propagations in every evaluated setting, while removing any individual criterion degrades all three search-cost metrics.
This indicates that C1, C2, and C3 contribute complementary supervision rather than being interchangeable.

The relative importance of individual criteria nevertheless varies across datasets and budgets.
C2 is particularly important for CA under the 5-call setting and for PS under the 3-call setting, where removing it causes the largest degradation.
By contrast, C1 has a stronger effect on SR with three calls and on PS with five calls.
Removing C3 also consistently weakens performance, although the magnitude of the degradation varies across settings.
Together, these results show that no single outcome comparison criterion explains the overall gain.
Combining completion, trajectory-level, and terminal evidence yields the most reliable solver-level improvements across different SAT distributions and intervention budgets.

\begin{center}
    \centering
    \includegraphics[width=0.95\textwidth]{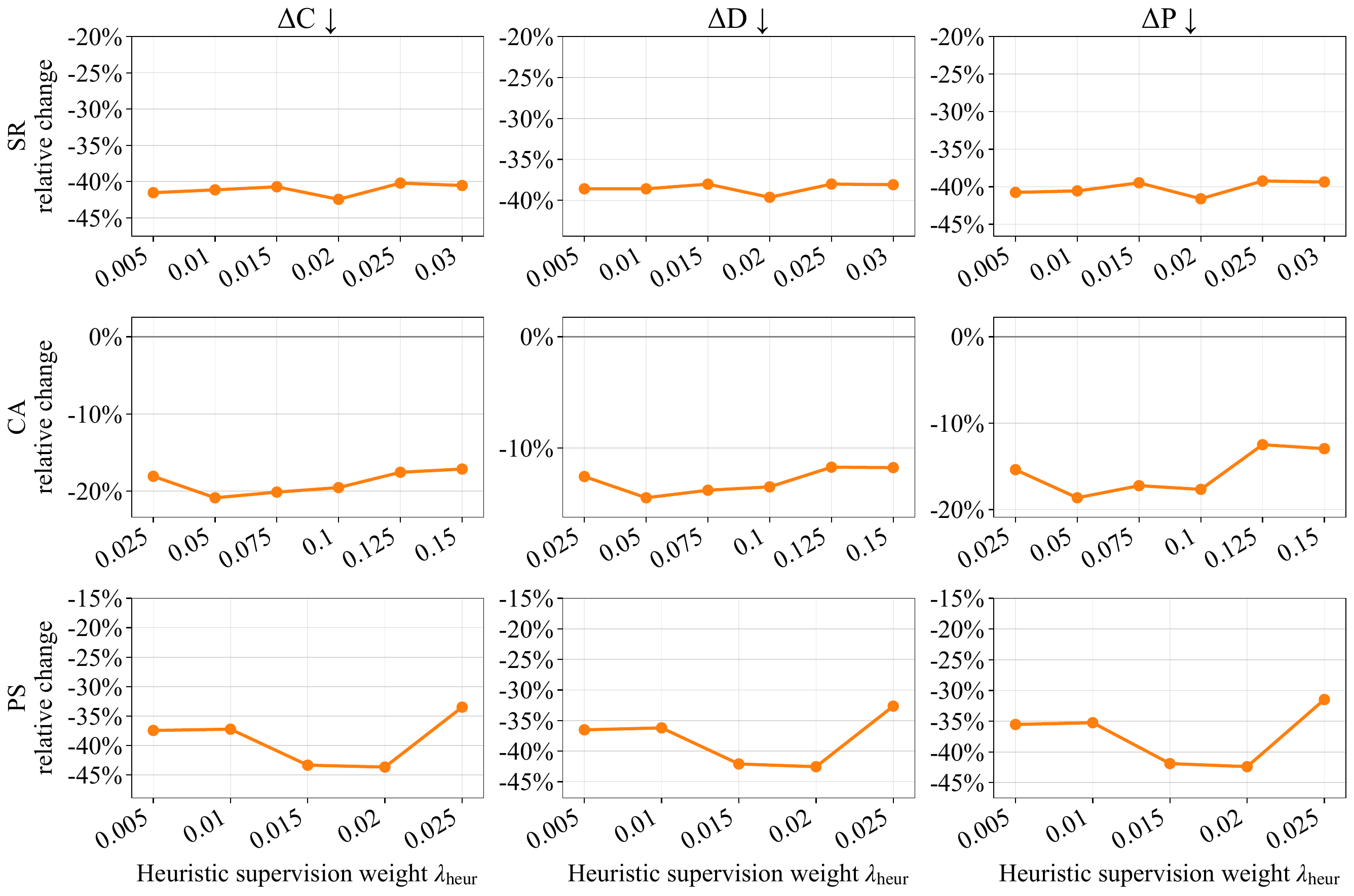}
    \captionof{figure}{
    Complete sensitivity analysis of the heuristic action supervision weight on SR, CA, and PS under the 5-call setting.
    The plots report relative changes in conflicts ($\Delta_C$), decisions ($\Delta_D$), and propagations ($\Delta_P$) with respect to native CaDiCaL; lower is better.
    }
    \label{fig:pan_weight_full}
\end{center}

\subsection{Sensitivity to Heuristic Action Supervision Weight}
\label{app:pan_weight_full}

Extending the propagation results in Figure~\ref{fig:pan_weight_main}, we
evaluate the sensitivity of non-dominated heuristic action supervision across
all three search-cost metrics under the 5-call setting.
The preference-guided ranking objective, model architecture, and evaluation
protocol remain fixed while $\lambda_{\mathrm{heur}}$ varies.
Figure~\ref{fig:pan_weight_full} reports the relative changes in conflicts,
decisions, and propagations.
The three solver-cost metrics exhibit broadly consistent trends across the tested weights.
SR remains relatively stable over the evaluated range.
CA shows comparable performance across moderate weights, with its strongest reductions generally occurring over $\lambda_{\mathrm{heur}}\in[0.05,0.1]$, while PS performs best over $\lambda_{\mathrm{heur}}\in[0.015,0.02]$.
Increasing the heuristic action supervision weight beyond these ranges does not consistently improve performance.
These results indicate that non-dominated heuristic action supervision performs best with a moderate weight.

\subsection{Generalization to Hard Instances}
\label{app:hard_generalization}

We further evaluate whether the models used in the main experiments generalize
to substantially harder SAT instances.
Specifically, we directly apply the same trained models used for the
main-experiment split to the corresponding Hard split, without retraining,
fine-tuning, or adaptation on any Hard instances.
Therefore, this experiment evaluates cross-difficulty generalization of the
learned branching policies rather than performance after additional training
on the target split.

The Hard instances are substantially larger than those in the main-experiment
split.
SR Hard and CA Hard
contain 200--400 variables, with average clause counts of 1616.47 and 4197.51,
respectively, while PS Hard contains 200--300 variables with an average of
1761.98 clauses.
We use the 3-call intervention setting for this generalization evaluation.
The results are reported in Table~\ref{tab:hard_generalization}.

\begin{table*}[!t]
\centering
\footnotesize
\caption{
Direct generalization of the same trained models used in the main experiments
to the corresponding Hard splits.
$\Delta_C$, $\Delta_D$, and $\Delta_P$ denote relative changes in conflicts,
decisions, and propagations, respectively.
Lower is better for $\Delta_C$, $\Delta_D$, $\Delta_P$, and MRPP;
higher is better for W(1\%).
}
\label{tab:hard_generalization}

\renewcommand{\arraystretch}{1.08}
\setlength{\tabcolsep}{8pt}

\begin{tabular}{llr@{\hspace{6pt}}r@{\hspace{6pt}}r@{\hspace{6pt}}r@{\hspace{6pt}}r}
\toprule
\textbf{Dataset}
& \textbf{Method}
& $\Delta_C \downarrow$
& $\Delta_D \downarrow$
& $\Delta_P \downarrow$
& \textbf{MRPP $\downarrow$}
& \textbf{W(1\%) $\uparrow$} \\
\midrule

\multirow{3}{*}{SR Hard}
& CaDiCaL
& 0.00\%
& 0.00\%
& 0.00\%
& 1.000
& 0.00\% \\

& \oursbase{}
& -25.95\%
& -25.28\%
& -25.69\%
& 0.704
& 78.10\% \\

& \ours{}
& \textbf{-28.24\%}
& \textbf{-27.44\%}
& \textbf{-27.87\%}
& \textbf{0.680}
& \textbf{81.40\%} \\

\midrule

\multirow{3}{*}{CA Hard}
& CaDiCaL
& 0.00\%
& 0.00\%
& 0.00\%
& \textbf{1.000}
& 0.00\% \\

& \oursbase{}
& -1.99\%
& -1.54\%
& -0.68\%
& 1.001
& 38.70\% \\

& \ours{}
& \textbf{-3.72\%}
& \textbf{-3.78\%}
& \textbf{-3.22\%}
& 1.001
& \textbf{41.00\%} \\

\midrule

\multirow{3}{*}{PS Hard}
& CaDiCaL
& 0.00\%
& 0.00\%
& 0.00\%
& 1.000
& 0.00\% \\

& \oursbase{}
& -3.29\%
& -3.37\%
& -3.00\%
& \textbf{0.461}
& 80.80\% \\

& \ours{}
& \textbf{-3.92\%}
& \textbf{-4.00\%}
& \textbf{-3.60\%}
& 0.482
& \textbf{81.00\%} \\

\bottomrule
\end{tabular}

\end{table*}

The same models used in the main experiments retain clear solver-level
improvements when directly transferred to the Hard splits.
On SR Hard, \ours{} reduces conflicts, decisions, and propagations by
28.24\%, 27.44\%, and 27.87\%, respectively, and also improves MRPP and
W(1\%) over \oursbase{}.
On CA Hard, \ours{} similarly achieves larger reductions in all three
aggregate search-cost metrics, although the overall gains are smaller than on
SR Hard.

On PS Hard, \ours{} improves the reductions in conflicts, decisions, and
propagations from 3.29\%, 3.37\%, and 3.00\% with \oursbase{} to
3.92\%, 4.00\%, and 3.60\%, respectively, while slightly improving W(1\%).
Its MRPP is slightly worse than that of \oursbase{}, showing that the
instance-level metrics do not improve uniformly.

Overall, the results show that the branching policies learned in the main
experiments remain effective when directly applied to substantially larger
Hard instances.
Since exactly the same trained models are used without any specific
training or adaptation, these results provide evidence that the trained
branching policies generalize beyond the instance scale observed in the
main-experiment split.

\subsection{Neural Inference Cost}
\label{app:runtime}

\paragraph{Measurement Protocol.}
We compare the neural inference cost of \ours{} with NeuroCore, GQSAT,
SATformer, ImitSAT, and RLAF.
For methods that provide online branching guidance, we report results under
matched 3-call and 5-call intervention budgets.
SATformer and RLAF instead perform a single initialization before search.
Inference time is measured by the model service and reported as average
accumulated milliseconds per instance.

\paragraph{Inference-Cost Comparison.}
Table~\ref{tab:runtime_full} reports the neural inference cost on SR, CA, and
PS.
These datasets contain relatively small instances that native CaDiCaL often
solves within milliseconds.
The neural inference cost therefore outweighs the runtime benefit obtained
from the reduced search cost on these instances.
The main results on SR, CA, and PS should consequently be interpreted as
improvements in search efficiency rather than end-to-end runtime acceleration.

Under matched intervention budgets, the inference cost of \ours{} remains
comparable to NeuroCore, GQSAT, and ImitSAT and is often lower than that of
the latter two methods.
SATformer and RLAF use one-shot initialization and do not provide repeated
guidance at dynamic solver states.
The online inference overhead of \ours{} is therefore comparable to existing
neural methods with similar deployment mechanisms.

The runtime tradeoff changes on the substantially harder SAT Competition
benchmarks analyzed in Appendix~\ref{app:sat_solver_combination}.
As shown in Table~\ref{tab:sat_solver_combination}, neural guidance reduces
PAR-2 on all three benchmark sets and solves additional SC-2024 and SC-2025
instances.
On the 752 instances solved by both configurations, it reduces total runtime
by 10.15\% after accounting for neural-call overhead.
These results show that the search-time reduction can outweigh neural
inference cost on sufficiently challenging SAT instances.

\begin{table*}[!t]
\centering
\footnotesize
\caption{
Neural inference cost of learning-based SAT branching methods.
All values are average accumulated milliseconds per instance.
Lower is better.
}
\label{tab:runtime_full}

\setlength{\tabcolsep}{6pt}

\begin{tabular}{lcrrr}
\toprule
\textbf{Method}
& \textbf{Budget}
& \multicolumn{3}{c}{\textbf{Inference (ms)}} \\
\cmidrule(lr){3-5}
& & \textbf{SR} & \textbf{CA} & \textbf{PS} \\
\midrule
NeuroCore
& 3-call
& 29.6 & 65.9 & 37.2 \\

NeuroCore
& 5-call
& 47.8 & 109.2 & 58.3 \\

GQSAT
& 3-call
& 66.1 & 153.8 & 84.3 \\

GQSAT
& 5-call
& 103.1 & 251.2 & 134.5 \\

SATformer
& one-shot
& 48.5 & 97.1 & 59.9 \\

ImitSAT
& 3-call
& 41.3 & 80.1 & 53.7 \\

ImitSAT
& 5-call
& 73.3 & 135.5 & 94.1 \\

RLAF
& one-shot
& 15.4 & 23.7 & 17.3 \\

\ours{}
& 3-call
& 42.1 & 60.4 & 46.7 \\

\ours{}
& 5-call
& 69.1 & 100.8 & 72.6 \\

\bottomrule
\end{tabular}
\end{table*}

\subsection{SAT Competition Evaluation}
\label{app:sat_solver_combination}

\paragraph{Training Instances and Data Split.}
This experiment uses an independently constructed training set derived from
the SAT Competition Main Track instances released between 2017 and 2022.
We partition the instances into training, validation, and test splits using an
80/10/10 ratio and a split seed of 42.
Before data collection, we retain formulas with at most 200{,}000 graph nodes
and 1{,}000{,}000 literal occurrences.
The graph contains two literal nodes for each variable and one node for each
clause.
No additional limits are imposed on the CNF file size or maximum clause
length.
The filtering process retains 956 formulas that satisfy these constraints and
pass DIMACS parsing.

\paragraph{State and Action Collection.}
The SAT Competition data collection uses a larger candidate set and a
different rollout budget from the G4SATBench setup.
Each retained state is restricted to 200{,}000 graph nodes and 1{,}000{,}000
literal occurrences.
At most 10 candidate variables are retained at each state, and both polarities
initiate separate rollouts.

State collection and action rollouts use CaDiCaL 3.0.0 with its default
solver configuration.
Each forced-action rollout has a timeout of 1{,}800 seconds.
No conflict, decision, or propagation limit is imposed.
The resulting collection contains 114{,}120 reliable forced-action rollouts.

After removing states without a supported action preference, the resulting
dataset contains 2{,}831 training states, 313 validation states, and 329 test
states.
The training, validation, and test splits contain 418{,}283, 45{,}312, and
48{,}272 directed preference pairs, respectively.

\paragraph{Solver Evaluation.}
Each of SC-2023, SC-2024, and SC-2025 contains 400 Main Track instances.
We evaluate native CaDiCaL and CaDiCaL guided by \ours{} on all instances from
the three benchmark sets.
Both configurations use CaDiCaL 3.0.0 with
\texttt{stabilizeonly=1} and \texttt{preprocesslight=0}.
All remaining solver options retain their default values.
Both CaDiCaL and neural inference run on the CPU in this evaluation.
Each solver run has a time limit of 5{,}000 seconds and uses the 3-call setting,
allowing at most three neural requests.
The two configurations otherwise use identical solver settings.

A neural request is permitted only when both the node and cell size checks
satisfy
$2|\mathcal{V}_t|+|\mathcal{C}_t|+|\mathcal{E}_t|\leq2{,}000{,}000$.
Here, $\mathcal{E}_t$ denotes the literal--clause incidence cells, which
correspond to literal occurrences.

We evaluate aggregate solving time using PAR-2,
\begin{equation}
    \mathrm{PAR2}
    =
    \frac{1}{N}
    \sum_{i=1}^{N}
    \begin{cases}
        T_i, & \text{if instance } i \text{ is solved},\\
        2T_{\max}, & \text{otherwise},
    \end{cases}
\end{equation}
where $T_i$ is the runtime of a solved instance and
$T_{\max}=5{,}000$ seconds.
Each timed-out instance therefore contributes 10{,}000 seconds.

\paragraph{Neural-Call Overhead.}
We report the macro-averaged NN wait ratio,
\begin{equation}
    \mathrm{NN\ Wait\ Ratio}
    =
    \frac{1}{N}
    \sum_{i=1}^{N}
    \frac{T_{\mathrm{NN\text{-}wait},i}}
         {T_{\mathrm{solve},i}},
\end{equation}
which assigns equal weight to each evaluated instance.
We also report the NN call latency,
\begin{equation}
    \mathrm{NN\ Call\ Latency}
    =
    \frac{\sum_i T_{\mathrm{NN\text{-}wait},i}}
         {\sum_i N_{\mathrm{RPC\text{-}attempts},i}}.
\end{equation}
The latency measures the end-to-end solver waiting time for each neural
request and covers communication, queueing, input processing, and model
inference.
Table~\ref{tab:sat_solver_call_cost} reports these measurements over all 400
instances in each benchmark set.

\begin{table}[t]
\centering
\footnotesize
\caption{
Neural-call overhead for the SAT Competition evaluation.
}
\label{tab:sat_solver_call_cost}
\setlength{\tabcolsep}{5pt}
\begin{tabular}{llrr}
\toprule
\textbf{Dataset} & \textbf{Solver} &
\textbf{NN Wait Ratio} &
\textbf{NN Call Latency (s)} \\
\midrule
\multirow{2}{*}{SC-2023}
& CaDiCaL (Native) & 0.00\% & -- \\
& CaDiCaL (\ours{}) & 11.99\% & 9.61 \\
\midrule
\multirow{2}{*}{SC-2024}
& CaDiCaL (Native) & 0.00\% & -- \\
& CaDiCaL (\ours{}) & 5.79\% & 8.10 \\
\midrule
\multirow{2}{*}{SC-2025}
& CaDiCaL (Native) & 0.00\% & -- \\
& CaDiCaL (\ours{}) & 7.92\% & 8.43 \\
\bottomrule
\end{tabular}
\end{table}

The NN wait ratios are 11.99\%, 5.79\%, and 7.92\% on SC-2023, SC-2024, and
SC-2025, respectively.
The corresponding NN call latencies are 9.61, 8.10, and 8.43 seconds.
These measurements cover both solved instances and runs that reach the
5{,}000-second time limit.

\paragraph{Analysis of Commonly Solved Instances.}
We define the commonly solved set as the instances for which both native and
\ours{}-guided CaDiCaL return SAT or UNSAT.
The two solvers produce identical SAT/UNSAT results on every instance in this
set.
Across the three benchmark sets, 752 instances are solved by both solvers,
16 are solved only by native CaDiCaL, 21 are solved only by the guided solver,
and 411 remain unsolved by both.
On SC-2023, both solvers solve 251 instances, but each uniquely solves four
instances.
The guided solver obtains net gains of two solved instances on SC-2024 and
three on SC-2025, resulting in five additional solved instances overall.

Among the 752 commonly solved instances, the guided solver is faster on 448
instances, corresponding to 59.57\%, while native CaDiCaL is faster on 304
instances.
The guided solver reduces the total runtime by 61{,}101.75 seconds, or
10.15\%, corresponding to an aggregate speedup of $1.113\times$.
We define the instance-level runtime ratio as
$R_i=T_{\mathrm{native},i}/T_{\mathrm{guided},i}$, where $R_i>1$ favors
\ours{}.
The median ratios are $1.051\times$, $1.128\times$, and $1.055\times$ on
SC-2023, SC-2024, and SC-2025, respectively.
The corresponding aggregate speedups are $1.080\times$, $1.073\times$, and
$1.193\times$.
SC-2024 shows the largest median improvement, whereas SC-2025 shows the
largest aggregate speedup.
This difference suggests that the SC-2025 gains are concentrated more heavily
on instances with longer solving times.

We also recompute the neural-call overhead on the commonly solved instances.
The NN wait ratios are 19.26\%, 8.76\%, and 12.62\% on SC-2023, SC-2024, and
SC-2025, while the corresponding NN call latencies are 12.33, 7.61, and
9.09 seconds.
Across all 752 instances, the NN wait ratio is 13.48\% and the NN call latency
is 9.67 seconds.
These wait ratios are higher than the full-set values because the full-set
averages contain many runs that reach the 5{,}000-second time limit.
A small number of neural calls accounts for only a small fraction of the total
runtime on these timed-out runs.
The commonly solved set therefore provides a more direct view of neural
overhead during successful solving.

Together, these results show that neural guidance maintains outcome
consistency, modestly expands solving coverage, and reduces end-to-end runtime
on commonly solved instances after accounting for neural-call overhead,
although the gains are not uniform across all instances.

\end{document}